\documentclass[10pt,twocolumn,letterpaper]{article}

\usepackage[pagenumbers]{cvpr} % To force page numbers, e.g. for an arXiv version

\definecolor{cvprblue}{rgb}{0.21,0.49,0.74}
\usepackage[pagebackref,breaklinks,colorlinks,allcolors=cvprblue]{hyperref}

\def\paperID{16265} % *** Enter the Paper ID here
\def\confName{CVPR}
\def\confYear{2025}

\title{ControlFace: Harnessing Facial Parametric Control for Face Rigging}

\author{
Wooseok Jang$^1$
\quad
Youngjun Hong$^2$
\quad
Geonho Cha$^2$
\quad
Seungryong Kim$^3$\\
$^1$Korea University \quad\quad $^2$NAVER Cloud \quad\quad $^3$KAIST
}

\usepackage{lipsum}  
\usepackage{amsmath}
\usepackage{float}
\usepackage{wrapfig}
\usepackage{array}
\usepackage{makecell}

\newcommand{\paragrapht}[1]{\vspace{-10pt}\paragraph{#1}}

\begin{document}
\twocolumn[{%
\renewcommand\twocolumn[1][]{#1}%
\maketitle

\begin{center}
    \centering
    \captionsetup{type=figure}
    \includegraphics[width=1\textwidth]{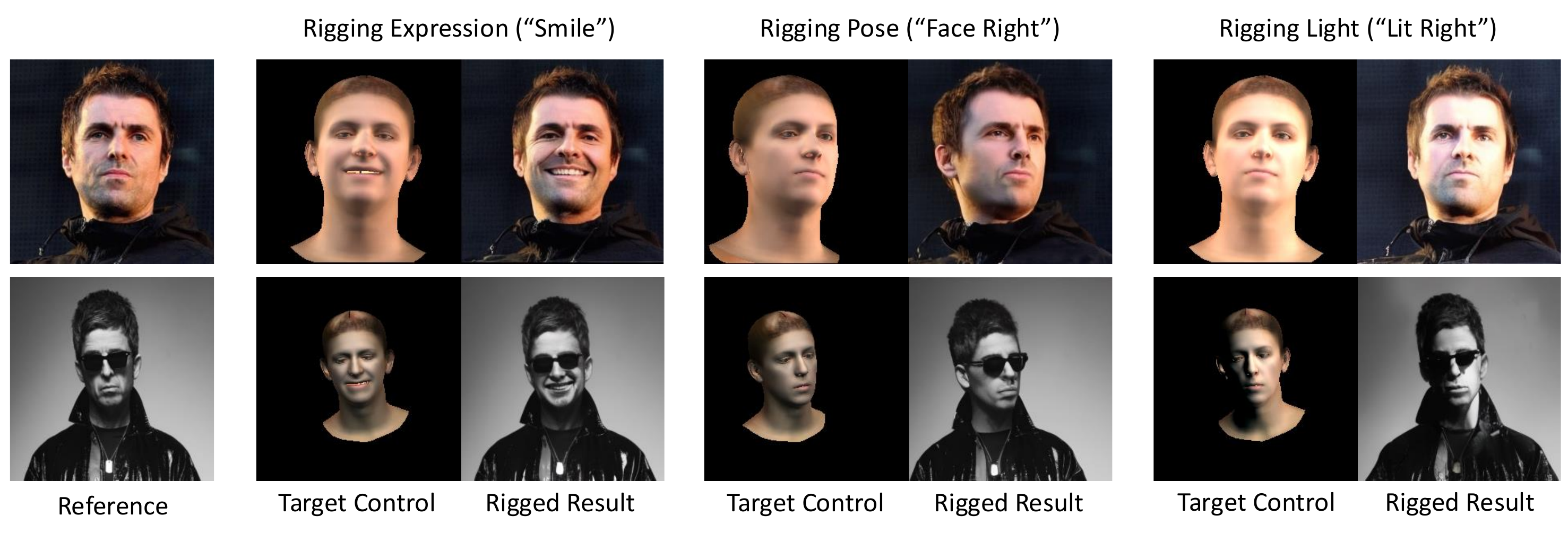}
    \vspace{-8mm}
    \captionof{figure}{Our \textbf{ControlFace} can edit the input face image using explicit facial parametric controls, generating realistic images without compromising the identity and other semantic details such as hairstyle.}
    \label{fig:teaser}
\end{center}%
%\vspace{-0.1cm}
}]

\begin{abstract}
Manipulation of facial images to meet specific controls such as pose, expression, and lighting, also known as face rigging, is a complex task in computer vision. Existing methods are limited by their reliance on image datasets, which necessitates individual-specific fine-tuning and limits their ability to retain fine-grained identity and semantic details, reducing practical usability. To overcome these limitations, we introduce ControlFace, a novel face rigging method conditioned on 3DMM renderings that enables flexible, high-fidelity control. We employ a dual-branch U-Nets: one, referred to as FaceNet, captures identity and fine details, while the other focuses on generation. To enhance control precision, the control mixer module encodes the correlated features between the target-aligned control and reference-aligned control, and a novel guidance method, reference control guidance, steers the generation process for better control adherence. By training on a facial video dataset, we fully utilize FaceNet’s rich representations while ensuring control adherence. Extensive experiments demonstrate ControlFace’s superior performance in identity preservation and control precision, highlighting its practicality. Please see the project website: \href{https://cvlab-kaist.github.io/ControlFace/}{https://cvlab-kaist.github.io/ControlFace/}.
\vspace{-10pt}
\end{abstract}    
\section{Introduction}

Manipulating input facial images has long been a fundamental task in computer vision and graphics~\cite{blanz2023morphable,thies2015real,liu20223d}. Given a face image, the goal of face rigging is to generate an image that corresponds to user controls—lighting or facial pose—without compromising the identity of the face. This task is often complicated by the need to preserve fine details like hairstyle, making it a challenging problem.

Recent advancements in generative models~\cite{liu20223d,song2020score,rombach2022high,esser2024scaling,ramesh2022hierarchical} have significantly impacted the development of face rigging techniques. Diffusion models, in particular, have demonstrated remarkable capabilities in producing high-quality, realistic facial images through their iterative refinement processes~\cite{song2020score,ho2020denoising}. These models have shown promise in generating realistic facial images. Leveraging 3D face reconstruction methods~\cite{feng2021learning,danvevcek2022emoca,zielonka2022towards}, enabling the estimation of 3D Morphable Model (3DMM) parameters—such as those of FLAME~\cite{li2017learning}—directly from an image, many works~\cite{tewari2020stylerig,ghosh2020gif,ding2023diffusionrig,jia2023discontrolface,liang2024caphuman,papantoniou2024arc2face} have utilized 3DMM parameters or renderings as target controls for face rigging. However, existing methods still face several limitations, including difficulties in handling real images~\cite{tewari2020stylerig,ghosh2020gif}, the requirement for fine-tuning on images of the same identity~\cite{ding2023diffusionrig,jia2023discontrolface}, and challenges in maintaining fine-grained details from reference images~\cite{liang2024caphuman,papantoniou2024arc2face}. Addressing these issues is crucial to enhance the practicality and robustness of face rigging approaches in real-world applications.

\begin{figure}[t] % Use "H" to force placement here (requires the float package)
    \centering
    \includegraphics[width=0.45\textwidth]{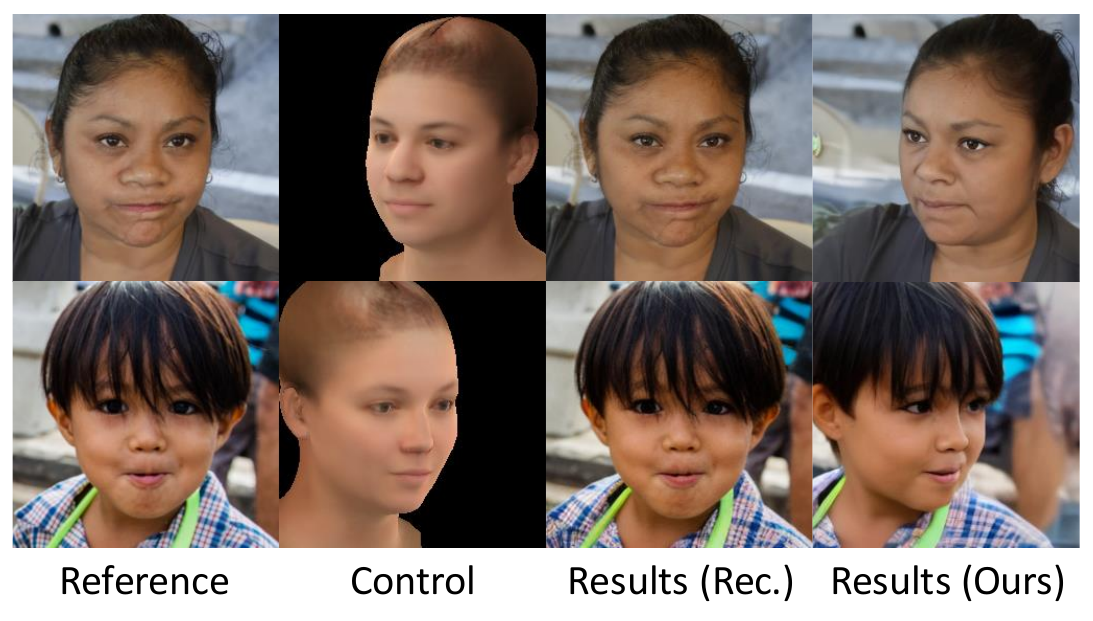} % Adjust width for one column
    \vspace{-2mm}
    \caption{\textbf{Limitations of reconstruction-based training.} We compare the results of our model trained on an image dataset~\cite{karras2019style} in a reconstruction setup and on a video dataset~\cite{zhu2022celebvhq} with paired samples created by randomly selecting two frames from each video. The results by reconstruction-based training often ignores the target control at inference.}
    \vspace{-10pt}
    \label{fig:recon}
\end{figure}

Notably, conventional diffusion-based face rigging methods~\cite{ding2023diffusionrig, jia2023discontrolface, liang2024caphuman, papantoniou2024arc2face} rely on single-image-per-individual datasets such as FFHQ~\cite{karras2019style}. Consequently, these methods are inevitably trained in a reconstruction manner, where only the reference image and its aligned control are available. Specifically, the reference image is embedded through a encoder and the model learns to reconstruct the input reference image with the embedded feature and the aligned control. However, if the feature from the reference image conveys too much information, reconstruction task becomes trivial, which often leads the model to ignore the control signal and simply replicate the reference image making rigging not possible as shown in Fig.~\ref{fig:recon}. Hence, to make the reconstruction non-trivial, they often encode the reference image into a single vector using a network such as pretrained face recognition model~\cite{deng2019arcface}. This compression into a single vector, however, may discard essential facial details failing to capture both identity and details simultaneously. This leads to additional individual specific fine-tuning~\cite{ding2023diffusionrig,jia2023discontrolface}. However, fine-tuning requires images with the same identity as the reference image which may be cumbersome to obtain.

In this paper, we introduce \textbf{ControlFace}, a novel face rigging method conditioned on 3DMM renderings, which effectively integrates both target control and reference control extracted from the reference image. Unlike previous methods~\cite{ghosh2020gif,liang2024caphuman,ding2023diffusionrig}, ControlFace enables flexible edits to lighting, shape, expression, and pose without requiring fine-tuning, while preserving both identity and semantic details of the input reference image. To achieve this, we use dual-branch U-Nets initialized from a large pretrained model: one, termed \textit{FaceNet}, which encodes the reference image to capture its identity and semantic details, and the second  \textit{denoising U-Net} that handles the generation. The rich representation encoded by the FaceNet is concatenated with the intermediate feature of the denoising U-Net and integrated through self-attention layer. For better control adherence we introduce the control mixer module (CMM), which embeds the correlated features between the target and reference controls. Additionally, we introduce a reference control guidance (RCG) mechanism that uses the reference control as a null condition label to improve grounding, inspired by Classifier-Free Guidance (CFG)~\cite{ho2022classifier}. ControlFace is trained on a facial video dataset~\cite{zhu2022celebvhq}, which provides paired reference images and target images aligned with the desired control. This training approach allows us to avoid reconstruction-based training and hence fully leverage FaceNet’s rich generative representation without ignoring the target control.

We conduct extensive experiments and ablation studies to rigorously evaluate the effectiveness of our proposed method. We demonstrate ControlFace's superior performance in identity preservation, semantic consistency, and control precision compared to existing methods, by evaluating on multiple unseen faces~\cite{karras2019style} We also show that ControlFace outperforms fine-tuning-based methods~\cite{ding2023diffusionrig} without the need for the cumbersome additional training. Ablation studies further highlight the distinct contributions of all the components in our model. These results validate our approach and illustrate its potential for real-world applications in face rigging.
\section{Related Work} 
% Our work is closely related to face editing with generative models and controllable generation, which are essential for identity and semantic preservation in diffusion models.
% \vspace{-10pt}
\paragraph{Generative Face Editing.} Since the introduction of GANs~\cite{goodfellow2020generative}, face generation has received significant attention. Early GAN-based methods~\cite{shen2020interfacegan, tov2021designing, harkonen2020ganspace} leveraged the disentangled latent space of StyleGAN~\cite{karras2019style} to manipulate facial attributes by finding specific directions in latent space. Other approaches~\cite{tewari2020stylerig, ghosh2020gif, liu20223d, shen2018faceid, deng2020disentangled} incorporated 3DMMs~\cite{blanz2023morphable, booth2018large} to conditionally control pose and expression through 3D structural guidance, improving realism and consistency.

Recently, diffusion models have emerged as a powerful alternative for generative face editing due to their stable training and high-quality output. For identity-preserving generation and manipulation, recent diffusion-based works either incorporate facial features directly~\cite{wang2024instantid, valevski2023face0, yan2023facestudio, liang2024caphuman, han2023generalist}, train networks to encode images~\cite{preechakul2022diffusion, ding2023diffusionrig, jia2023discontrolface, ye2023ip}, or leverage pretrained diffusion U-Nets for rich representations~\cite{hu2024animate, xu2023magicanimate, zhu2024champ}. 

\paragrapht{Controllable Generation for Diffusion Models.} Methods for conditioning diffusion models with spatial controls initially relied on concatenation~\cite{saharia2022image, saharia2022palette, rombach2022high}. The advent of large-scale pretrained text-to-image models~\cite{rombach2022high, esser2024scaling} has leveraged sophisticated text embeddings~\cite{radford2021learningtransferablevisualmodels, raffel2020exploring} processed through advanced attention mechanisms, driving research into finer control over the generation process~\cite{zhang2023adding, peng2024controlnext, mou2024t2i, li2404controlnet++}. Recently, ControlNet~\cite{zhang2023adding} has enabled precise control over attributes like depth, edges, and poses by integrating timestep-specific control features into the diffusion U-Net. In face editing, approaches~\cite{ding2023diffusionrig, liang2024caphuman, jia2023discontrolface, han2023generalist} have adopted similar architectures with 3DMMs to allow explicit facial control. However, these methods often struggle to encode fine details from the reference image. To overcome this limitation, we employ dual-branch structure: one for encoding the reference facial image and the other for generating manipulated image.

\section{Preliminaries}

\paragraph{3D Morphable Face Models (3DMMs).}
3DMMs~\cite{blanz2023morphable,li2017learning,booth2018large} are facial parametric models designed to accurately represent facial pose, shape, and expression using a compact latent. FLAME~\cite{li2017learning} which is one of the widely used 3DMM, takes shape, pose, and expression as inputs and outputs a face mesh. 
In this work, we leverage DECA~\cite{feng2021learning} model which is capable of predicting the FLAME~\cite{li2017learning} parameters together with the orthographic camera parameters, albedo, and lighting in spherical harmonics from a facial image. With these parameters we can apply Lambertian reflectance to acquire surface normals, albedo, and Lambertian rendering. These renderings offer pixel-aligned control, providing both geometric and texture information that suitable for face rigging. Hence, we leverage these renderings as our control.

\paragrapht{Latent Diffusion Models (LDM).}
LDM~\cite{rombach2022high} is a generative model that progressively learns to denoise random Gaussian noise in the latent space of VAE~\cite{kingma2013auto}. During training, an input image 
$X$ is encoded into a latent representation $z$ via a VAE encoder. This latent $z$ is then perturbed with Gaussian noise $\epsilon$ through a predefined diffusion forward process at a randomly selected timestep $t$, producing the noisy latent $z_t$. The model’s core objective is to predict $\epsilon$, effectively learning to reverse the noise. The training objective can be written as:
\begin{equation}
\mathcal{L}_{LDM} = \mathbb{E}_{z, t, \epsilon\sim\mathcal{N}(0,1)}[\|\epsilon_{\theta}(z_t;t)-\epsilon\|^2_2].
\label{eq:ldm_loss}
\end{equation}

\paragrapht{Classifier-Free Guidance (CFG).}
To enhance conditional generation of diffusion model during inference, CFG~\cite{ho2022classifier} reformulates the predicted noise steering the generation process away from the unconditional distribution and towards the conditional distribution. This allows us to sample from a sharper conditional distribution, enhancing the condition fidelity. The modified prediction can be formulated as follows:
\begin{equation} \hat{\epsilon}_{\theta}(\cdot, C) = \epsilon_{\theta}(\cdot, \varnothing) + w \left( \epsilon_{\theta}(\cdot, C) - \epsilon_{\theta}(\cdot, \varnothing) \right), \label{eq:cfg} \end{equation}
where $\varnothing$ denotes an unconditional label such as empty text, $C$ denotes the conditional input and $w$ is a guidance scale. Hence, $\epsilon_{\theta}(\cdot, \varnothing)$ is the unconditional prediction and $\epsilon_{\theta}(\cdot, C)$ is the conditional prediction. Many diffusion models, including text-to-image models~\cite{rombach2022high,esser2024scaling,betker2023improving,ramesh2022hierarchical}, employ this technique to achieve high-quality generation.

\begin{figure*}[t]
    \centering
    \includegraphics[width=\textwidth]{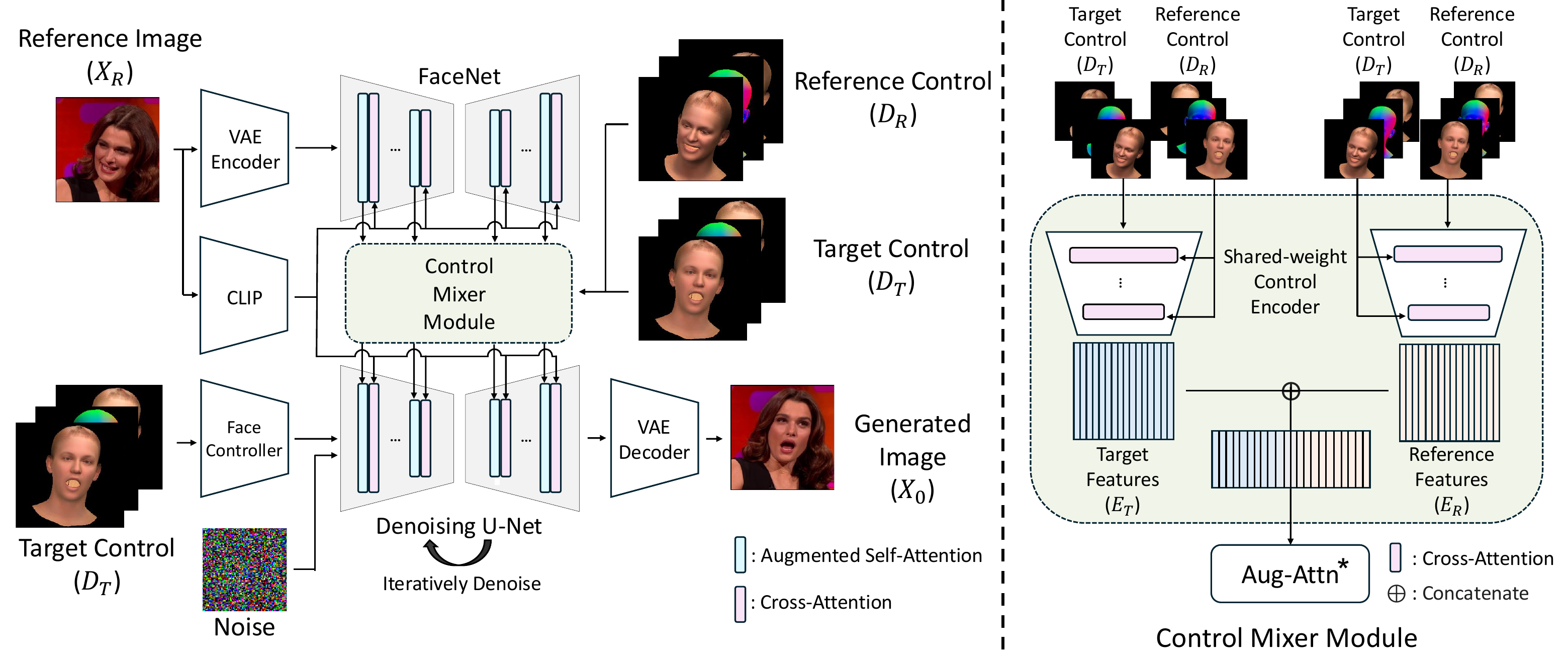}
    \vspace{-6mm}
    \caption{\textbf{Overall Architecture.} ControlFace encodes the reference image $X_{R}$ into the FaceNet and CLIP image encoder for identity and semantic preservation. For face control, the target control $D_{T}$ is incorporated into the denoising U-Net through face controller. To enhance the control adherence, the correlated feature between reference control $D_{R}$ and target control $D_{T}$ is acquired from the proposed control mixer module. }
    \label{fig:Architecture}
    \vspace{-10pt}
\end{figure*}
\section{Method}
\subsection{Overview}
Given a facial reference image $X_{R} \in \mathbb{R}^{H \times W \times 3}$ with height $H$ and weight $W$, our objective is to generate a target facial image $X_{T}$ that retains the identity and detailed attributes of $X_{R}$, while aligning with an image-aligned target control $D_{T} \in \mathbb{R}^{H \times W \times 9}$, which is defined by 3DMM renderings~\cite{li2017learning,feng2021learning} consisting of surface normal, albedo, and Lambertian renderings. 

We employ dual branch U-Nets: the proposed FaceNet to capture the local and fine-grained appearance of $X_{R}$ and a denoising U-Net for generation. FaceNet's detailed features are injected into the denoising U-Net through the augmented self-attention layer. Additionally,  we utilize CLIP~\cite{radford2021learningtransferablevisualmodels} image encoder to provide coarse visual features to both U-Nets. To guide manipulation by control signal, we present a lightweight encoder called face controller, which embedds $D_{T}$ directly into the denoising U-Net. However, as it only encodes $D_{T}$, the model cannot fully map the relation between $X_{T}$ and $X_{R}$, limiting control adherence. Hence, we propose control mixer module (CMM) with multiple cross-attention layers to embed correlated features between $D_{T}$ and $D_{R}$ which is then fused into the augmented self-attention layer to guide the model's attention effectively. $D_{R}$ is another 3DMM renderings extracted from the reference image. Fig.~\ref{fig:Architecture} illustrates our overall architecture. For further control adherence, we employ reference control guidance (RCG) during inference. Each component is explained in details in the following sections. 

\subsection{Acquiring Paired Quadruplets for Face Rigging}
As shown in Fig.~\ref{fig:recon}, training our model with image datasets where $X_{R}=X_{T}$ like previous methods~\cite{ding2023diffusionrig,jia2023discontrolface}  may lead the model to replicate $X_{R}$ and ignore $D_{T}$. Hence, we leverage facial video dataset, CelebV-HQ~\cite{zhu2022celebvhq}, which contains 35,666 facial video clips involving 15,653 identities with various actions. From each video, we randomly select two frames, assigning one as $X_{R}$ and the other as $X_{T}$. We then extract image-aligned surface normals, albedo, and Lambertian rendering from each frames. The extracted renderings from the reference image serve as the reference control $D_{R}$, while those from the target image serve as the target control $D_{T}$.
 Consequently, our dataset consists of a set of quadruplets consisting of $\{X_{R},X_{T},D_{R},D_{T}\}$.

\subsection{Injecting Appearance of Reference Image}
To acquire strong image representations that preserve identity and fine details during generation, the reference image is encoded using both the CLIP~\cite{radford2021learningtransferablevisualmodels} image encoder and the proposed FaceNet. The CLIP image embeddings, which capture high-level semantic information, are integrated into both denoising U-Net and FaceNet through cross-attention layers. Simultaneously, FaceNet encodes the reference facial image to extract detailed features which are then incorporated into the denoising U-Net through augmented self-attention mechanism~\cite{hu2024animate,xu2023magicanimate,seo2024genwarp,shi2024instadrag}. Since FaceNet and the denoising U-Net share the same architecture, they integrate seamlessly in self-attention by concatenating keys and values from both U-Nets and applying attention operation with queries from the denoising U-Net. Let $Q$, $K$, $V\in\mathbb{R}^{l\text{x}d}$ denote query, key, value from the denoising U-Net where $l$ and $d$ denote length and dimensionality. Also, let $K^\mathrm{face}$, $V^\mathrm{face}\in\mathbb{R}^{l\text{x}d}$ represent key, and value from the FaceNet, respectively. Then the augmented self-attention $\mathrm{Aug\text{-}Attn}$ is formulated as follows: 
\begin{equation}
\mathrm{Aug\text{-}Attn}
=\mathrm{Softmax}\big(\frac{Q[K,K^\mathrm{face}]^T}{\sqrt{d}}\big)[V,V^\mathrm{face}],
\label{eq:self-attn}
\end{equation}
where $[\cdot,\cdot]$ denotes concatenation, and $\mathrm{Softmax}(\cdot)$ means a softmax operation.

\subsection{Incorporating Target Controls}
\paragraph{Face Controller.} To manipulate the reference image, we directly encode the target control \( D_{T} \) using a CNN-based shallow network which we denote face controller. The encoded features from the face controller are added to the first layer of the denoising U-Net. We have tested alternative diffusion conditioning methods, like ControlNet~\cite{zhang2023adding} and ControlNeXt~\cite{peng2024controlnext}, but led to worse performance with increased computational complexity. The ablation study on the conditioning methods can be found in the Section~\ref{sec:5_exp}.

\paragrapht{Control Mixer Module.}
CMM outputs correlated embeddings for both controls $D_{R}$ and $D_{T}$, denoted $E_{R}$, and $E_{T}$, respectively. CMM encodes these controls using two shared-weight control encoders. The control encoder consists of multiple convolution layers along with the cross-attention layers which is capable of modeling the interaction between the two control effectively. Specifically, when one control is input, the other serves as a conditioning input, passed through a convolution layer to match the resolution with the input.

The structure of CMM is illustrated in Fig.~\ref{fig:Architecture}. The resulting embeddings $E_{R}$ and $E_{T}$ are in four different resolutions corresponding to the four different resolutions of U-Net activation maps. These embeddings are fused with features from the two U-Nets in the augmented self-attention layer. We modify Eq.~\ref{eq:self-attn} by adding $E_{R}$ and $E_{T}$ to the corresponding keys and values and denote it as $\mathrm{Aug\text{-}Attn^*}$. This can provide guidance on where the model should attend to. The modified augmented self-attention can be formulated as:
\begin{align}
    & \mathrm{Aug\text{-}Attn^*} = \nonumber &&  \\
    & \mathrm{Softmax} \big(\frac{ (Q + E_{T}) [ K + E_{T},K^\mathrm{face}+ E_{R} ]^T}{\sqrt{d}} \big)[V,V^\mathrm{face}].
\end{align}

\subsection{Training Objective}
Since we perform the generation process in the latent space of a VAE~\cite{kingma2013auto}, we encode $X_{R}$ and $X_{T}$ via the VAE encoder and acquire latents $z_{R}$ and $z_{T}$, respectively. Our model $\epsilon_\theta(\cdot)$ takes $z_{T}$,$z_{R}$, $D_{T}$, and $D_{R}$ as input and learns to predict random Gaussian Noise $\epsilon$ which is used to perturb $z_{R}$. Similar to Eq.~\ref{eq:ldm_loss}, our training objective is as follows:
\begin{align}
&\mathcal{L}
= \mathbb{E}_{z, t, \epsilon\sim\mathcal{N}(0,1)}[\|\epsilon_{\theta}(z_{T,t};t,z_{R},D_{T},D_{R})-\epsilon\|^2_2],
\end{align}
% \begin{align}
% &\mathcal{L}
% = \mathbb{E}_{z, t, \epsilon\sim\mathcal{N}(0,1)}[\|\epsilon_{\theta}(z_{R,t};t,z_{R},[E_{R},E_{T}],\mathcal{E}(D_{T})-\epsilon\|^2_2],
% \end{align}
where $z_{R,t}$ is the perturbed $z_{R}$ with predefined diffusion forward schedule with randomly selected timestep $t$.

\begin{figure}[t] % Use "H" to force placement here (requires the float package)
    \centering
    \includegraphics[width=0.47\textwidth]{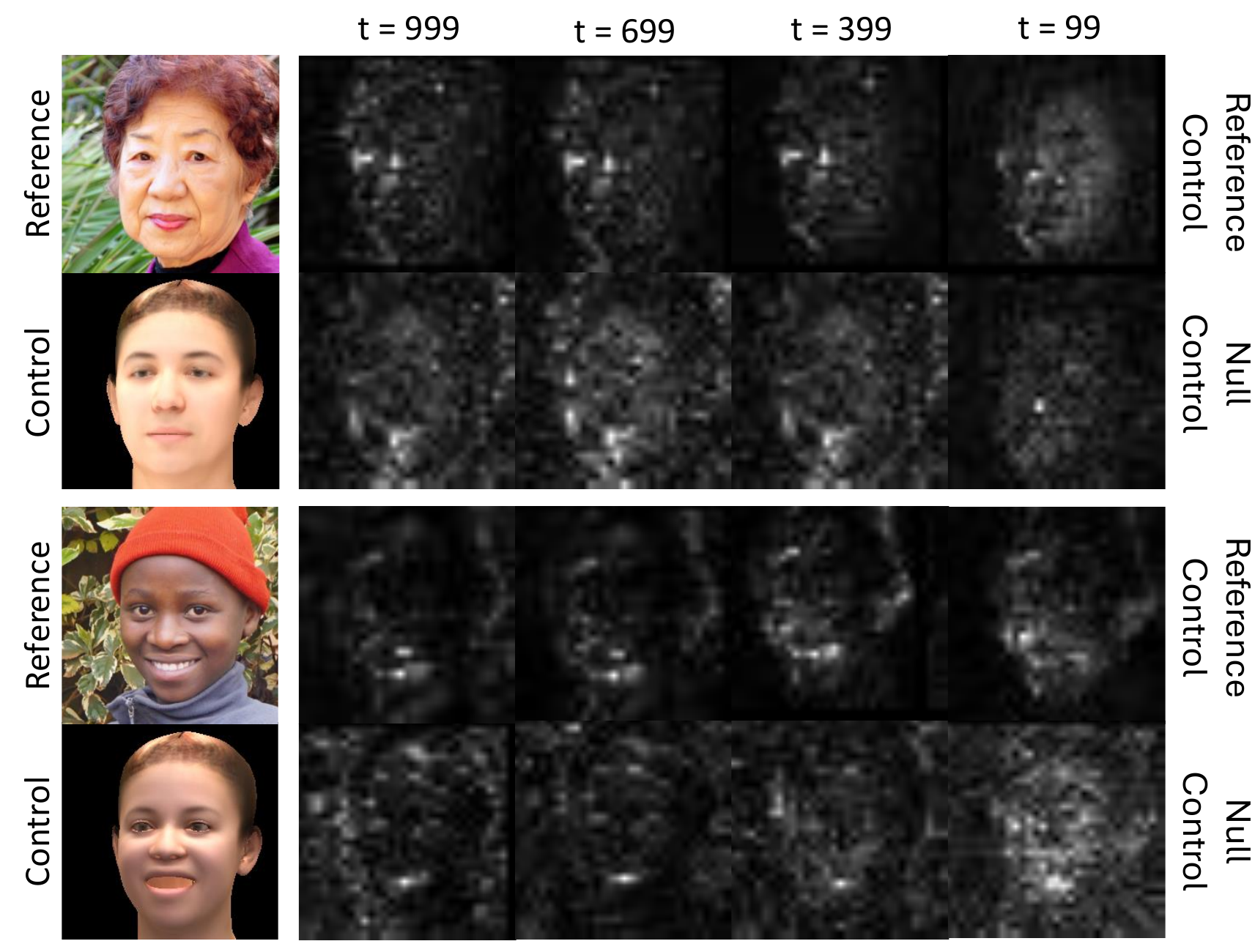} % Adjust width for one column
    \vspace{-2mm}
    \caption{\textbf{Visualization of Reference Control Guidance.} We visualize the deltas, $ \epsilon_{\theta}(\cdot, D_{T}) - \epsilon_{\theta}(\cdot, \varnothing)$ and $ \epsilon_{\theta}(\cdot, D_{T}) - \epsilon_{\theta}(\cdot, D_{R})$, which corresponds to CFG~\cite{ho2022classifier} applied to face controller input and RCG, respectively, across different timesteps $t$. The first and third row display RCG deltas whereas second and fourth row show the CFG deltas. The former shows noisy deltas over all the timesteps.}
    \vspace{-10pt}
    \label{fig:ref vis}
\end{figure}

\subsection{Reference Control Guidance}
At inference, we further enhance the control adherence with a novel guidance technique. To achieve this, we first naively apply CFG to face controller input. This can be formulated as follows:
\begin{equation} \hat{\epsilon}_{\theta}(\cdot, D_{T}) = \epsilon_{\theta}(\cdot, \varnothing) + w \big( \epsilon_{\theta}(\cdot, D_{T}) - \epsilon_{\theta}(\cdot, \varnothing) \big), \end{equation}
where $\varnothing$ is a null input for the face controller, similar to Eq.~\ref{eq:cfg}. However, Tab.~\ref{tab:ablation} shows that the improvement is negligible compared to the result without any guidance. We suspect that this is because the null condition cannot provide a good grounding for where to deviate away from. Therefore, our proposed guidance method, RCG, utilizes $D_{R}$ instead of null conditioning for better grounding. We define it as follows: 
\begin{equation} \hat{\epsilon}_{\theta}(\cdot, D_{T}) = \epsilon_{\theta}(\cdot, D_{R}) + w \big( \epsilon_{\theta}(\cdot, D_{T}) - \epsilon_{\theta}(\cdot, D_{R}) \big). 
\label{eq:RCG}
\end{equation}

In Fig.~\ref{fig:ref vis}, we visualize each difference, $ \epsilon_{\theta}(\cdot, D_{T}) - \epsilon_{\theta}(\cdot, \varnothing)$ and $ \epsilon_{\theta}(\cdot, D_{T}) - \epsilon_{\theta}(\cdot, D_{R})$, on various timesteps. Although the difference of both guidances tends to focus on facial area on lower timesteps, the former is noisy over all timesteps which can compromise the generation process. On the other hand, the difference of RCG shows relatively clean face-aligned estimates over all timesteps. Hence, we leverage Eq.~\ref{eq:RCG} for generating our output $X_0$.
\begin{figure*}[t]
    \centering
    \includegraphics[width=\textwidth]{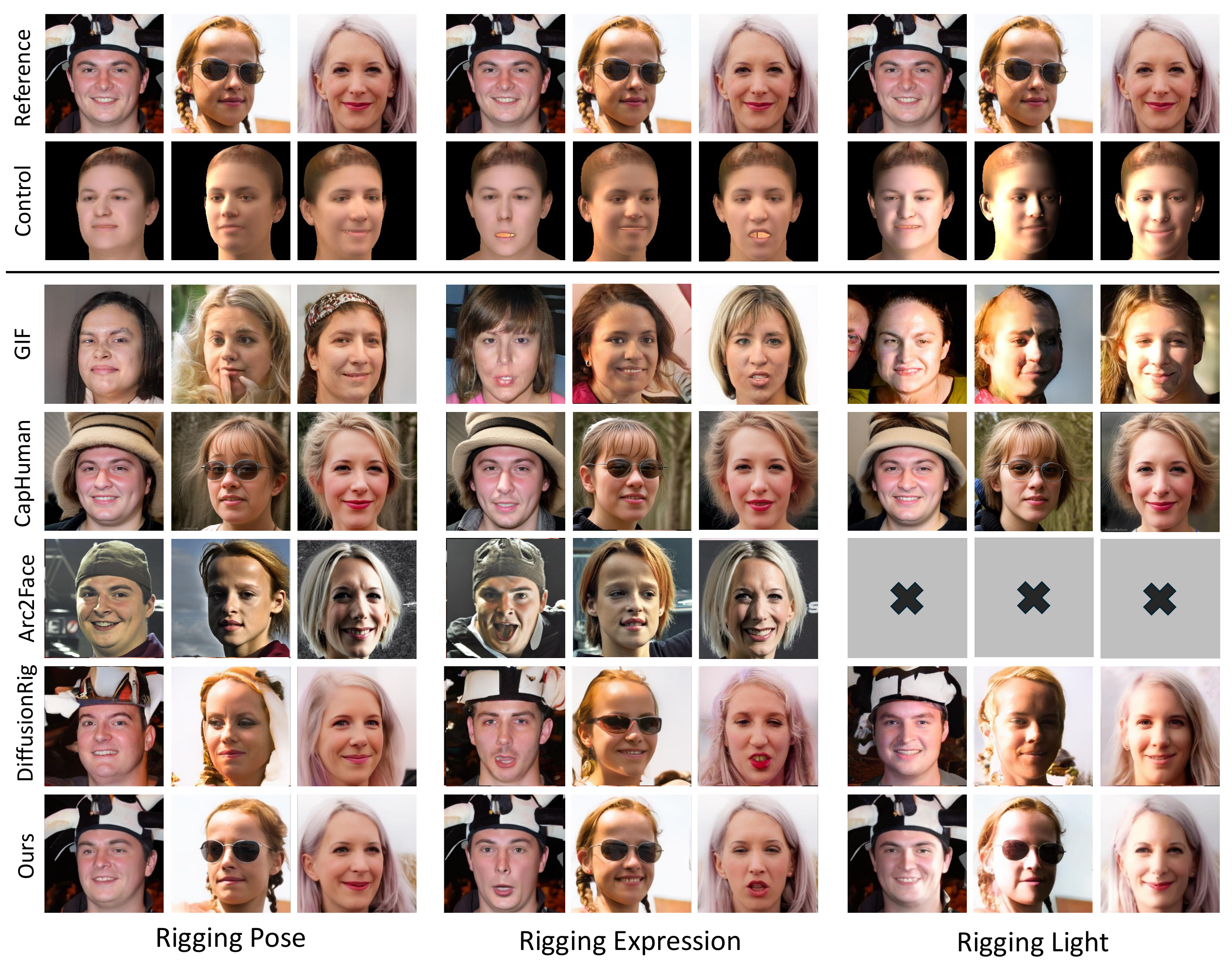}
    \vspace{-6mm}
    \caption{\textbf{Qualitative Results.} We compare the results of rigging pose, expression, and light with four different baselines~\cite{ghosh2020gif,liang2024caphuman,papantoniou2024arc2face,ding2023diffusionrig}. The reference images are from the FFHQ~\cite{karras2019style} dataset. Compared to the baselines ControlFace aligns with the target control, maintaining the identity and details in the reference image.}
    \label{fig:main_qul}
    \vspace{-10pt}
\end{figure*}

\section{Experiments}

% In this section, we first present details on our implementation. Then we provide qualitative comparisons with the baselines. We also provide quantitative results on control adherence and identity preservation along with a user study. Finally, ablation study highlights the impact of each model component.

\subsection{Implementation Details}
We train our model on CelebV-HQ~\cite{zhu2022celebvhq} dataset, resizing all frames to 256×256 pixels. We extract 3DMM renderings also of shape 256×256 from each frame leveraging DECA~\cite{feng2021learning}. The weights of the FaceNet and the denoising U-Net are first initialized from a variant of pretrained Stable Diffusion v1.5~\cite{rombach2022high}, which is a large text-to-image model fine-tuned to accept CLIP~\cite{radford2021learningtransferablevisualmodels} image embeddings as input. We present more implementation details in the Appendix~\ref{ap:implementation}.

\subsection{Comparison}
\paragraph{Baselines.}
We compare our method with four publicly available 3DMM-controllable face generation methods: GIF~\cite{ghosh2020gif}, CapHuman~\cite{liang2024caphuman}, Arc2Face~\cite{papantoniou2024arc2face}, and DiffusionRig~\cite{ding2023diffusionrig}. GIF is a 3DMM-conditional GAN-based approach utilizing the StyleGAN~\cite{karras2019style} generator, but it does not take a reference image as input. Other methods are diffusion-based. CapHuman and Arc2Face utilize a face recognition model~\cite{deng2019arcface} feature for identity preservation, whereas DiffusionRig employs ResNet~\cite{he2016deep} to encode the reference image features. CapHuman also accepts text input for additional control; therefore, we fix the input text to ``a photo of a person" in our experiments to ensure a fair comparison.

We evaluate ControlFace and the baselines on the FFHQ~\cite{karras2019style} dataset, which consists of over $70,000$ high-quality facial images. Notably, all baselines except CapHuman were trained on FFHQ, whereas our method was not. Additionally, ControlFace is trained on CelebV-HQ~\cite{zhu2022celebvhq}, which contains approximately $15,000$ unique identities—fewer than FFHQ. This demonstrates our method's robustness and generalizable capability.

\begin{figure}[t] % Use "H" to force placement here (requires the float package)
    \centering
    \includegraphics[width=0.47\textwidth]{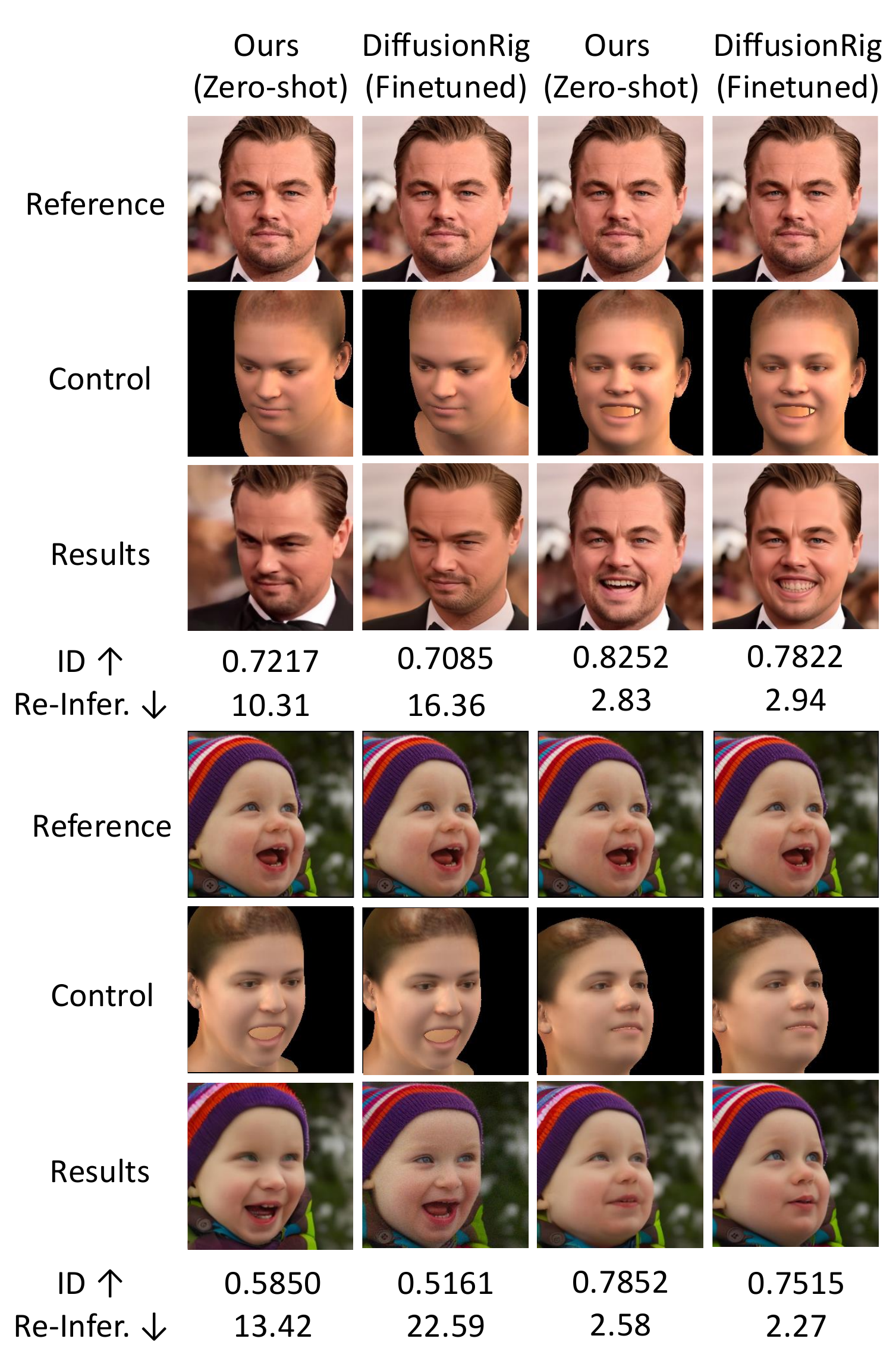} % Adjust width for one column
    \vspace{-2mm}
    \caption{\textbf{Comparison with Fine-tuning Method.} We compare the results with the fine-tuned DiffusionRig~\cite{ding2023diffusionrig} on two reference images. We show the generated results along with the identity similarity (ID)~\cite{deng2019arcface} and DECA~\cite{feng2021learning} re-inference error.}
    \vspace{-10pt}
    \label{fig:fine_qual}
\end{figure}
\paragrapht{Qualitative Comparison.}
We compare our generated results with the baselines on three identities from the FFHQ dataset. Fig.~\ref{fig:main_qul} shows the results. We present rigged outputs with variations in three attributes—pose, expression, and lighting—for all methods except Arc2Face, as it does not support lighting control. For visualization, we display only the Lambertian rendering of the control.

While all methods produce results aligned with the control parameters, GIF fails to preserve identity as it does not incorporate any image-related features. CapHuman and Arc2Face maintain the identity but lose semantic details like hairstyle and background since they utilize only facial features. DiffusionRig shows good performance in identity and detail preservation; however, the identity tends to change when the control differ significantly from those of the reference image. Compared to all baselines, our method exhibits superior performance in both identity and detail preservation while adhering to the target controls. We present more generated results on FFHQ dataset in the Appendix~\ref{ap:qual ffhq}.

Moreover, since DiffusionRig originally requires additional training on a personal album for optimal performance, we fine-tune it on the reference image. As shown in Fig.~\ref{fig:fine_qual}, while fine-tuning improves DiffusionRig's identity and detail preservation, it still struggles with identity shifts under large pose changes. In contrast, our method maintains consistent identity even under large pose variations, demonstrating robustness and fidelity in generated results.

\paragrapht{Quantitative Comparison.}
We measure the DECA re-inference error using the same settings as previous works~\cite{ghosh2020gif,ding2023diffusionrig,jia2023discontrolface} to evaluate fidelity to the target controls. Specifically, we generate 1,024 images by varying shape, lighting, expression, and pose parameters. For each generated image, we use DECA to re-inference the 3DMM parameters. We then calculate the Root Mean Square Error (RMSE) between the FLAME mesh rendered with original control parameters and the newly inferred ones. For lighting, we compute the RMSE directly on the spherical harmonics coefficients. Tab.~\ref{tab:quantitative results} shows that our method achieves the best control adherence on average.

We evaluate image quality using three metrics: identity preservation (ID) score via cosine similarity of features from a face recognition model~\cite{deng2019arcface}, Fréchet Inception Distance (FID)~\cite{heusel2017gans} for image quality, and Learned Perceptual Image Patch Similarity (LPIPS)~\cite{zhang2018unreasonable} for semantic consistency. These metrics are computed on the same set of 4096 images used for DECA re-inference error evaluation. ID and LPIPS are not calculated for GIF, as it does not use a reference image. Results in Tab.~\ref{tab:quantitative results face} show our method achieves the lowest FID and LPIPS, indicating high image quality and semantic consistency. While Arc2Face scores highest in identity preservation due to explicit facial feature conditioning, its high LPIPS score indicates a lack of background preservation. In contrast, our method not only preserves identity but also maintains semantic consistency with the reference image, offering a balanced performance across all evaluated metrics. Also, for a fair comparison, we train DiffusionRig on CelebV-HQ and apply the same evaluation metrics which is illustrated in the Appendix~\ref{ap:video}.
\begin{table}[t]
    \centering
    \caption{\textbf{Quantitative Results on Control Adherence.} We compare the DECA~\cite{feng2021learning} re-inference errors following previous work~\cite{ghosh2020gif} with four different baselines~\cite{ghosh2020gif,liang2024caphuman,papantoniou2024arc2face,ding2023diffusionrig}. Our method, ControlFace, shows the best control adherence on average.}
    \vspace{-2mm}
    \scalebox{0.90}{
    \begin{tabular}{l@{\hspace{5pt}}c@{\hspace{5pt}}c@{\hspace{5pt}}c@{\hspace{5pt}}c|@{\hspace{5pt}}c@{\hspace{5pt}}}
    \hline
    &Light$\downarrow$&  Shape $\downarrow$ &  Exp. $\downarrow$ &Pose $\downarrow$ &  Avg. $\downarrow$  \\
    \hline
 GIF \cite{ghosh2020gif}                                                
 & 17.04& 2.29  & 8.16 & 8.17 & 8.91 \\
 CapHuman~\cite{liang2024caphuman}
      &  15.16& 2.65& 6.68 & 19.03&10.40\\
Arc2Face~\cite{deng2019arcface}
     &  - & 3.13& 13.83 & 15.37 & 10.77\\
 DiffusionRig~\cite{ding2023diffusionrig}             
 & 6.31 & \textbf{2.11} & 5.58& \textbf{6.26}&5.06\\
 \hline
 \textbf{ControlFace (Ours)}& \textbf{3.75}&2.56& \textbf{5.43}& 7.67&\textbf{4.85}\\
     \hline
    \end{tabular}
    }
    \label{tab:quantitative results}
\end{table}
\begin{table}[t]
    \centering
    \caption{\textbf{Quantitative Results on Identity Preservation and Image Quality.} We evaluate identity similarity (ID)~\cite{deng2019arcface} and LPIPS~\cite{zhang2018unreasonable} to assess appearance preservation of the reference image, and FID~\cite{heusel2017gans} to measure overall image quality. }
    \vspace{-2mm}
    
    \begin{tabular}{l@{\hspace{5pt}}c@{\hspace{5pt}}c@{\hspace{5pt}}c}
    \hline
    &ID$\uparrow$&  FID $\downarrow$ &  LPIPS $\downarrow$\\
    \hline
 GIF \cite{ghosh2020gif}                                                
        & - & 24.48 & -  \\
 CapHuman \cite{liang2024caphuman}
        &  0.4356&59.78 & 0.4556 \\
Arc2Face \cite{deng2019arcface}
        & \textbf{0.7825} & 17.82& 0.5253  \\
 DiffusionRig \cite{ding2023diffusionrig}             
        & 0.2042 & 23.05 & 0.3758\\
 \hline
 \textbf{ControlFace (Ours)}& 0.7586&\textbf{15.50}& \textbf{0.1429}\\
     \hline
    \end{tabular}
    \label{tab:quantitative results face}
\end{table}
\begin{table}[t]
    \centering
    \caption{\textbf{User Study.} We ask the participants to evaluate the model on condition alignment (SC) and perceptual quality (PQ). Our model achieves the highest score on all metrics. }
    \vspace{-2mm}
    \begin{tabular}{l@{\hspace{5pt}}c@{\hspace{5pt}}c|@{\hspace{5pt}}c}
    \hline
    &SC$\uparrow$&  PQ $\uparrow$ & OV $\uparrow$\\
    \hline
 CapHuman~\cite{liang2024caphuman}                                                
        & 0.4086 & 0.7019 & 0.4545  \\
 DiffusionRig~\cite{ding2023diffusionrig}
        & 0.3533& 0.5072 & 0.3473 \\
\hline
 \textbf{ControlFace (Ours)}
        & \textbf{0.8750} & \textbf{0.8605}& \textbf{0.8486}  \\
     \hline
    \end{tabular}
    \label{tab:user study}
    \vspace{-10pt}
\end{table}

We conduct a user study for additional evaluation, following ImagenHub standards~\cite{ku2024imagenhub}. Participants rate each model on semantic consistency (SC), measuring alignment between the condition and the generated image, and perceptual quality (PQ), assessing image quality. We compare our model with CapHuman and DiffusionRig, both of which use reference images and lighting as conditions. Tab.\ref{tab:user study} presents SC, PQ, and the overall score (OV), which is the geometric mean of SC and PQ demonstrating the final score of the model. Our model achieves the highest scores across all metrics. Further details of the user study is available in the Appendix~\ref{ap:user}.

\subsection{Result on Out-of-Domain images}
\begin{figure}[t] % Use "H" to force placement here (requires the float package)
    \centering
    \includegraphics[width=0.47\textwidth]{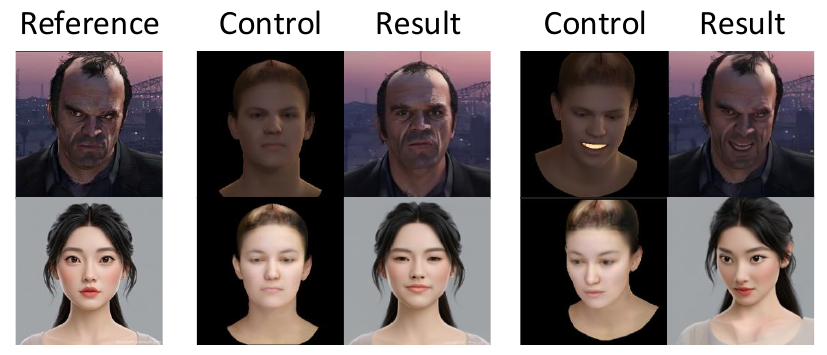} % Adjust width for one column
    \vspace{-2mm}
    \caption{\textbf{Results on Out-of-domain Data.} We test ControlFace on animation-like faces to illustrate the robustness and generalizability. }
    \label{fig:ood_qual}
\end{figure}

To further evaluate the zero-shot generalization capability of our model, we test ControlFace on out-of-domain reference images that significantly differ from the real face data leveraged during training. Specifically, we employ animation-like facial images as reference images. As shown in  Fig.~\ref{fig:ood_qual}, our model successfully generates rigged results for these animation faces, demonstrating that it can effectively handle outside the scope of its training data. We present more examples in the Appendix.~\ref{ap:qual ood}.
\vspace{-5pt}
\begin{figure}[t] % Use "H" to force placement here (requires the float package)
    \centering
    \includegraphics[width=0.40\textwidth]{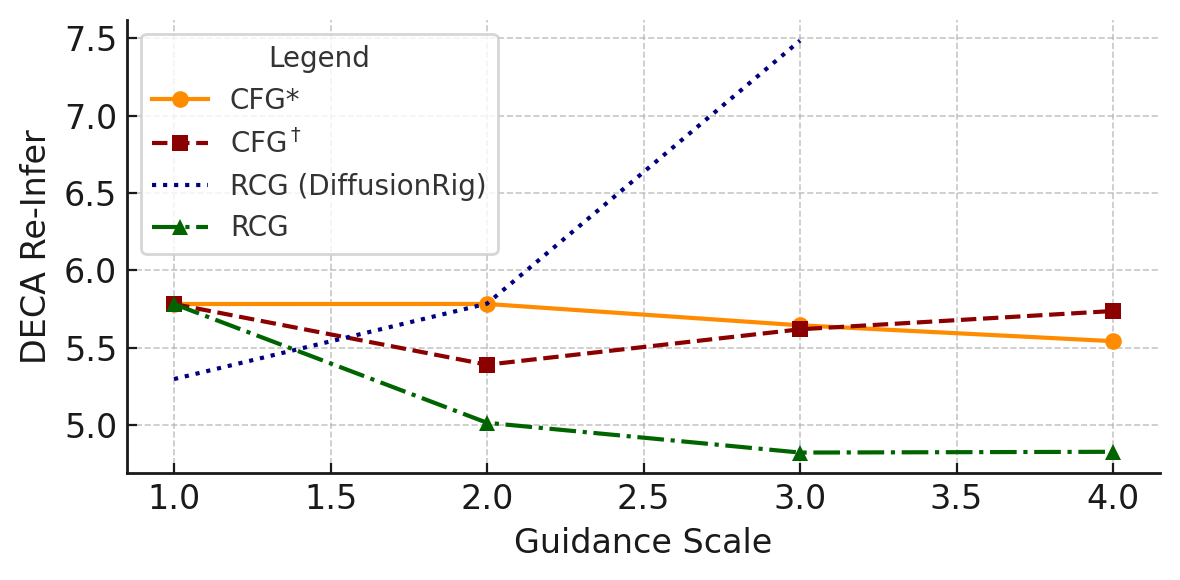} % Adjust width for one column
    \vspace{-2mm}
    \caption{\textbf{Ablation on Guidance Scale.} We plot the DECA~\cite{feng2021learning} re-inference error for guidance scales $w$ from 1.0 to 4.0. Increasing $w$ improves errors for RCG, $\text{CFG}^*$, and $\text{CFG}^\dagger$ to some extents, but greatly worsens for DiffusionRig~\cite{ding2023diffusionrig}.}
    \vspace{-10pt}
    \label{fig:ablation guidance}
\end{figure}
\subsection{Ablation Studies}
\label{sec:5_exp}
\paragraph{Scaling Reference Control Guidance.}
We examine the effect of guidance scales $w$ on control adherence. From here, $\text{CFG}^*$ and $\text{CFG}^\dagger$ refer to CFG applied to the face controller and CMM, respectively, both incorporating null inputs. On the other hand, CFG represents applying a null input to the CLIP encoder, following the approach used in most text-to-image models~\cite{rombach2022high,esser2024scaling,ramesh2022hierarchical}. Since RCG requires no null condition training, we also apply it to DiffusionRig, evaluating on 128 randomly selected FFHQ reference images. Fig.~\ref{fig:ablation guidance} shows that increasing the scale improves DECA re-inference error for RCG, $\text{CFG}^*$, and $\text{CFG}^\dagger$ to varying extents, but significantly reduces performance for DiffusionRig. We suspect this is due to DiffusionRig’s struggle to maintain identity consistency under different controls, causing conflicting identities and poor alignment when using both target and reference controls.

\paragrapht{Ablation on Different Conditioning Modules.}
We attach different kinds of widely used conditioning modules~\cite{zhang2023adding,peng2024controlnext} to see their effects on control adherence. On top of FaceNet archetecture, without CMM, we attach various conditioning modules, applying CFG during inference. Tab.~\ref{tab:cont abl results} compare face controller with ControlNet~\cite{zhang2023adding} and ControlNeXt~\cite{peng2024controlnext} showing the number of parameters and DECA re-inference error. Face controller performs the best while having the least number of trainable parameters.

\paragrapht{Ablation on Architecture.}
Starting with FaceNet and Face Controller, we sequentially add CMM, CFG~\cite{ho2022classifier}, $\text{CFG}^*$, and our proposed RCG, with results in Tab.~\ref{tab:ablation}. Incorporating CMM improves alignment and image quality but slightly reduces identity similarity compared to FaceNet. This occurs because enhanced control adherence requires adjustments to attributes like expression, pose, or lighting, which may inadvertently alter identity-related features. Consequently, FaceNet, with poorer alignment, achieves the highest identity preservation. Our proposed RCG method compensates for the marginal decrease in ID score by achieving better control adherence and image quality than FaceNet. Also, RCG exhibits better control alignment among other guidance methods without sacrificing the identity. We additionally provide ablation study on different inputs and architecture for CCM in the Appendix~\ref{ap:landmark}.

\subsection{Limitation}
ControlFace relies on DECA~\cite{feng2021learning} for acquiring 3DMM renderings which may not capture the finer facial details needed for optimal performance. Adopting a more advanced 3D face estimation model~\cite{danvevcek2022emoca,zielonka2022towards} could enhance the model’s overall performance. Additionally, ControlFace is currently trained on a single facial video dataset~\cite{zhu2022celebvhq}, limiting its exposure to a variety conditions and identities. Incorporating additional video datasets, as well as multiview facial datasets, could further improve the model’s generalizability.

\begin{table}[t]
    \centering
    \caption{\textbf{Conditioning Module Ablation.} We plot parameter counts and DECA~\cite{feng2021learning} re-inference errors for various conditioning modules.}
    \vspace{-2mm}

    \begin{tabular}{l@{\hspace{5pt}}c@{\hspace{5pt}}c@{\hspace{5pt}}}
    \hline
    & \# of params. &  Re-Infer.   \\
    \hline
ControlNet~\cite{zhang2023adding}
      &  $\mathrel{\sim}$360M &8.37\\
ControlNeXt~\cite{peng2024controlnext}
     & $\mathrel{\sim}$3M & 8.06\\
     \hline
  \textbf{Face controller (Ours)}
 & \textbf{$\mathrel{\sim}$1M} & \textbf{7.87} \\
     \hline
    \end{tabular}
    \label{tab:cont abl results}
\end{table}

\begin{table}[t]
    \centering
    \caption{\textbf{Architecture Ablation.} We plot DECA~\cite{feng2021learning} re-inference error, identity similarity score~\cite{deng2019arcface}, FID~\cite{heusel2017gans}, and LPIPS~\cite{zhang2018unreasonable} for each architecture.}
    \vspace{-2mm}
    \scalebox{0.88}{
    \begin{tabular}{l@{\hspace{5pt}}|c@{\hspace{5pt}}|c@{\hspace{5pt}}c@{\hspace{5pt}}c@{\hspace{5pt}}c@{\hspace{5pt}}}
    \hline
    &Component  &Re-Infer.$\downarrow$&  ID $\uparrow$ &  FID $\downarrow$ &LPIPS \\
    \hline
    (I)& FaceNet &  7.13& \textbf{0.8234}& 32.45 & \textbf{0.1321}\\
    \hline
    (II) &(I) + CMM &  5.78 & 0.7520& \textbf{15.35} & 0.1384 \\
    \hline
    (III)& (II) + CFG  &  5.77 & 0.7586 & 15.50 & 0.1429 \\
    (IV) &(II) + $\text{CFG}^*$ &  5.75 & 0.7308 & 15.43 & 0.1400 \\
    (V) &\textbf{(II) + RCG (Ours)} & \textbf{4.85} & 0.7586 & 15.50& 0.1429 \\
    \hline
    \end{tabular}
    }
    \label{tab:ablation}
    \vspace{-10pt}
\end{table}

\section{Conclusion}
In this paper, we introduce ControlFace, a face rigging method that addresses limitations arising from reliance on image datasets, such as loss of fine details and need for fine-tuning. Our dual-branch U-Net is capable of capturing the identity and intricate details of the reference image. To improve control adherence, we leverage CMM which learns embedding the relationship between the target and reference controls, and RCG which utilizes reference control for better grounding. Training on a facial video dataset improved enhanced identity and detail preservation. Experiments and ablations confirm ControlFace’s superior performance in identity consistency, control adherence, and image quality, demonstrating its practicality for real-world applications.
\vspace{-20pt}
{
    \small
    \bibliographystyle{ieeenat_fullname}
    \bibliography{main}
}

% \maketitlesupplementary
\onecolumn

% Manually add the supplementary title
\begin{center}
    \textbf{\Large ControlFace: Harnessing Facial Parametric Control for Face Rigging}
    \vspace{0.5em} \\
    \textbf{\large - Supplementary Material -}
\end{center}
\vspace{-5pt}
% \vspace{5pt}
\appendix
\setcounter{page}{1}
% In the supplementary material, we outline implementation details, provide user study settings, present ablation studies, and include an analysis related to RCG, along with additional generation results from our method, ControlFace.
In the supplementary material, we include additional details on the implementation and user study settings. We also present further ablation studies and provide an in-depth analysis of RCG. Finally, we showcase additional generation results from our method, ControlFace.
\section{More on Implementation Details}
\label{ap:implementation}

\begin{wraptable}{r}{0.4\textwidth}\vspace{-45pt}

    \begin{center}
    \caption{\textbf{Number of parameters.} We report parameter counts for each proposed component.}
    \vspace{-2mm}

    \begin{tabular}{l@{\hspace{5pt}}c@{\hspace{5pt}}c@{\hspace{5pt}}}
    \hline
    & \# of parameters.   \\
    \hline
FaceNet
      &  $\mathrel{\sim}$850M \\
Face Controller
     & $\mathrel{\sim}$3M\\
CMM
 & $\mathrel{\sim}$30M \\
     \hline
     \vspace{-30pt}
    \end{tabular}
    \label{tab:param}
    \end{center}
    \end{wraptable}

\paragraph{Training Details.}
The model was trained for 300,000 steps on 8 NVIDIA V100 GPUs. Each GPU processed a batch size of 4, resulting in a total effective batch size of 32. We trained all of our components except for VAE encoder~\cite{kingma2013auto} and CLIP image encoder~\cite{radford2021learningtransferablevisualmodels}. Tab.~\ref{tab:param} shows the number of trainable parameters for each component of ControlFace. We employed 8bit-Adam optimizer~\cite{dettmers20218} for memory efficiency and a constant learning rate of 0.00001 throughout the training process. The entire training procedure took approximately three days to complete.

\paragrapht{Inference Details.}
For all the results in the qualitative and quantitative comparison in the main paper, ControlFace, CapHuman~\cite{liang2024caphuman}, and Arc2Face~\cite{papantoniou2024arc2face} utilize DDIM~\cite{song2020denoising} scheduler with 50 sampling steps. The guidance scale for ControlFace was set to 4 while CapHuman and Arc2Face was set to 3.5 which is the default value. For DiffusionRig~\cite{ding2023diffusionrig}, we utilize DDPM~\cite{ho2020denoising} scheduler with 250 sampling steps. The computational cost of RCG is equivalent to that of CFG~\cite{ho2022classifier}, as they both double the batch size to process both the conditional and unconditional paths simultaneously.

\begin{wrapfigure}{r}{0.4\textwidth}\vspace{-56pt}
\begin{center}
 \includegraphics[width=1.\linewidth]{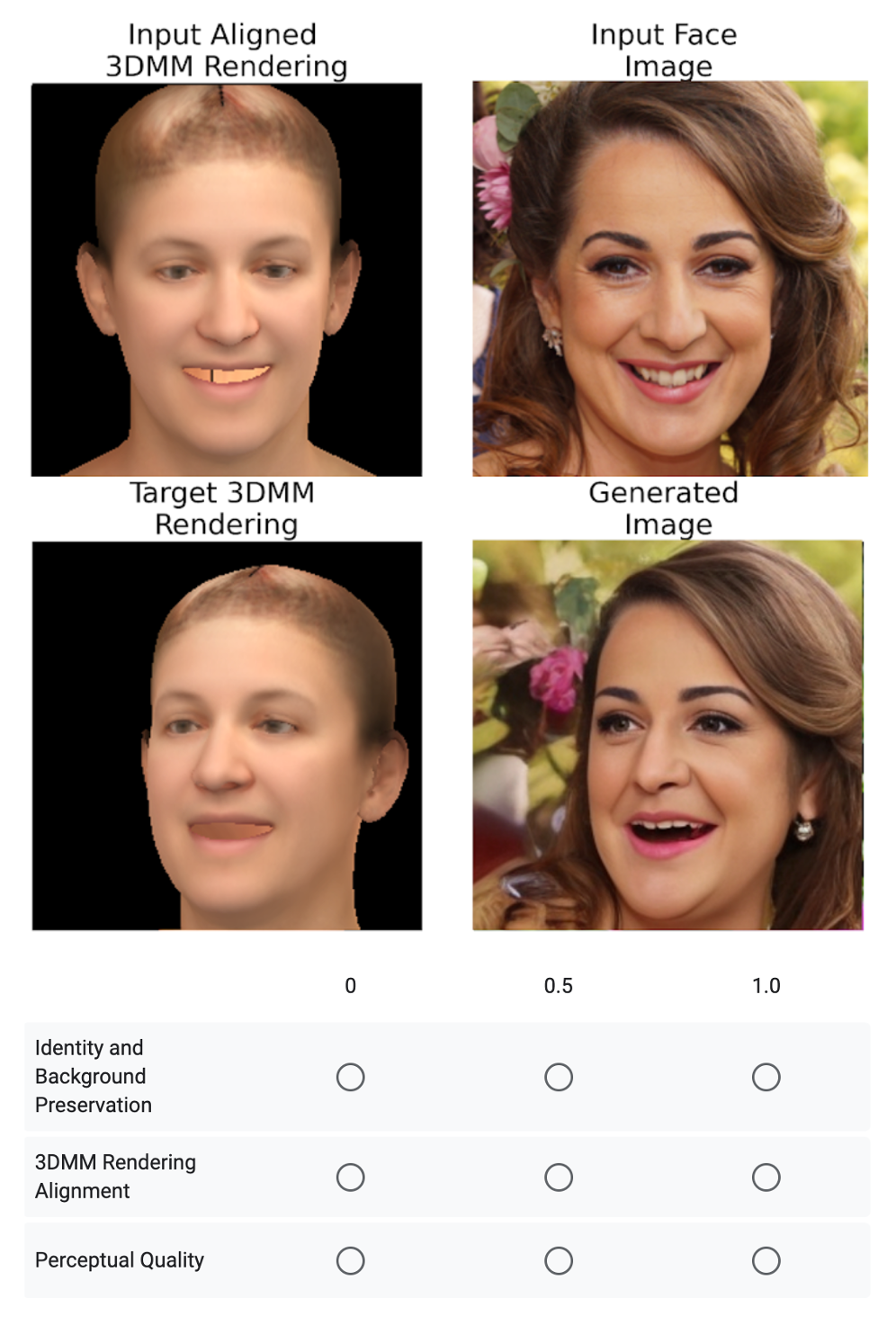}

\vspace{-10pt}
\caption{\textbf{Example of user study.} For each generated result, the participants were asked three questions. }
\label{fig:user}
\end{center}
\vspace{-40pt}
\end{wrapfigure}

\section{User Study Settings}
\label{ap:user}

We provide detailed information about the user study. Following the methodology of ImagenHub~\cite{ku2024imagenhub}, participants evaluated each model based on two criteria: (1) how well the generated image aligns with the input condition (SC) and (2) the overall quality (PQ) of the generated image. Since ControlFace takes both a reference image and 3DMM renderings as input conditions, participants assessed its performance separately for each condition, with the SC defined as the lower score between the two evaluations. Eight participants were recruited and divided into two groups, each group evaluating different examples consisting of 52 generated images per model. Fig.~\ref{fig:user} illustrates a sample page from our user study.
\section{Additional Experiments}
\subsection{Training DiffuisonRig~\cite{ding2023diffusionrig} on Video Dataset}
For a fair comparison, we train two versions of DiffusionRig~\cite{ding2023diffusionrig} on CelebV-HQ~\cite{zhu2022celebvhq}: a reconstruction version where the reference and target images are identical and a paired version where they differ. Both versions are evaluated using the same protocol outlined in the main paper. Tab.~\ref{tab:vid} shows a significant drop in performance for the newly trained models compared to the original implementation. This is due to the limited identities in CelebV-HQ and the ResNet in DiffusionRig struggling to encode the rich information in the video data, highlighting the need for our architectural design choices.
\label{ap:video}

\subsection{Model with Only CLIP Embedding}
\begin{wraptable}{r}{0.4\textwidth}\vspace{-50pt}
\begin{center}
\caption{\textbf{Quantitative results of model with only CLIP~\cite{radford2021learningtransferablevisualmodels} image encoder.} We compare the result of CLIP image encoder and FaceNet.}
\vspace{-5pt}
    \scalebox{0.85}{
    \begin{tabular}{l@{\hspace{5pt}}|c@{\hspace{5pt}}c@{\hspace{5pt}}c@{\hspace{5pt}}c@{\hspace{5pt}}c@{\hspace{5pt}}}
    \hline
      &Re-Infer.$\downarrow$&  ID $\uparrow$ &  FID $\downarrow$ &LPIPS $\downarrow$ \\
    \hline
    CLIP~\cite{radford2021learningtransferablevisualmodels}
     & 8.96 & 0.1623 & \textbf{29.77} & 0.4496 \\
    + FaceNet
     &  \textbf{7.13} & \textbf{0.8234} & 32.45 & \textbf{0.1321} \\
    \hline
    \end{tabular}
    \label{tab:clip}
}    
\vspace{-20pt}
\end{center}
\end{wraptable}

\begin{wrapfigure}{r}{0.4\textwidth}\vspace{-48pt}
\begin{center}
 \includegraphics[width=1.\linewidth]{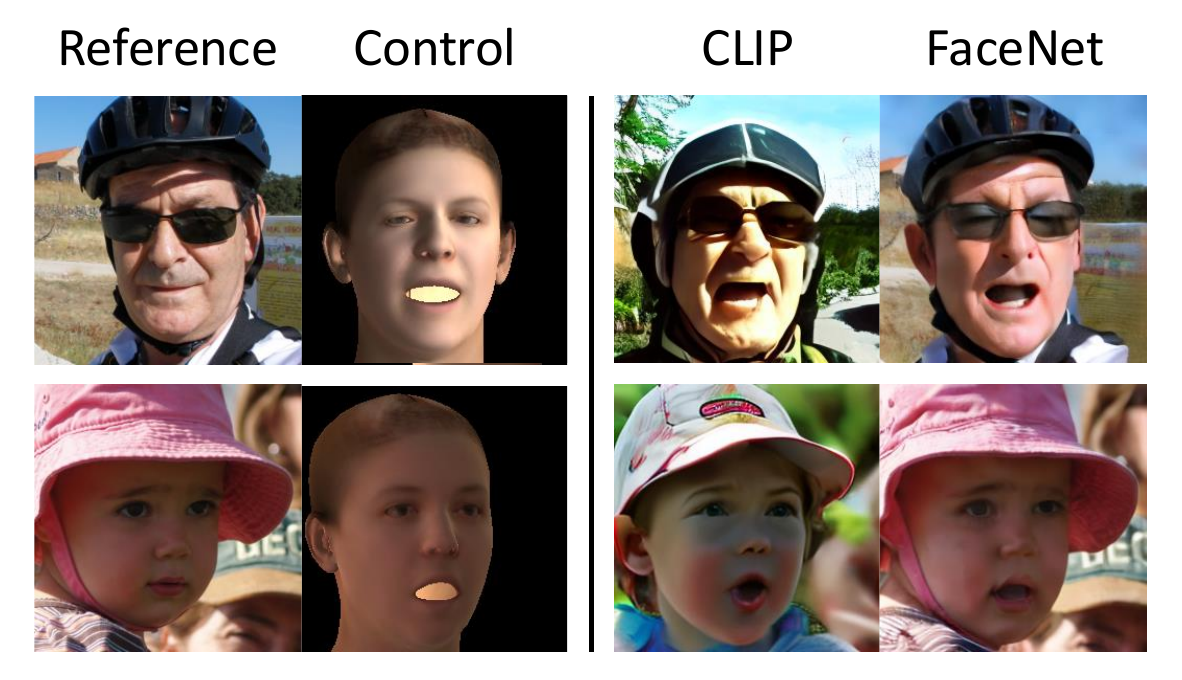}

\vspace{-5pt}
\caption{\textbf{Qualitative results of model with only CLIP~\cite{radford2021learningtransferablevisualmodels} image encoder.} We train the denoising U-Net with only CLIP image encoder attached and compare the results with the FaceNet. }
\label{fig:clip}
\end{center}
\vspace{-20pt}
\end{wrapfigure}

We train the denoising U-Net with only the CLIP~\cite{radford2021learningtransferablevisualmodels} image encoder attached and evaluate its performance using the same metrics described in the main paper. The results, as presented in Tab.~\ref{tab:clip}, reveal that the CLIP image encoder alone struggles to fully encode the reference image, resulting in poor performance. This limitation emphasizes the importance of our proposed FaceNet, which is specifically designed to capture and preserve the fine details and identity present in the reference image, ensuring improved results. Although FaceNet achieves lower FID~\cite{heusel2017gans}, we show in Fig.~\ref{fig:clip} that the model which utilizes CLIP embedding often generates over-saturated results, which degrades the image quality.

\subsection{More ablation on CMM}
\label{ap:landmark}

We trained two additional models using different inputs to the CMM to identify the most suitable method to embed the relationship between the reference image and the target image. First, we trained a model using facial landmarks extracted from FLAME~\cite{feng2021learning}. Specifically, we convert the landmarks into a map by assigning an integer value to the index corresponding to each landmark. Next, we employed Lambertian rendering, one of three types of renderings used as control in our method, which offers more information compared to the other two. 

\begin{figure}[t]
\begin{center}
 \includegraphics[width=1\linewidth]{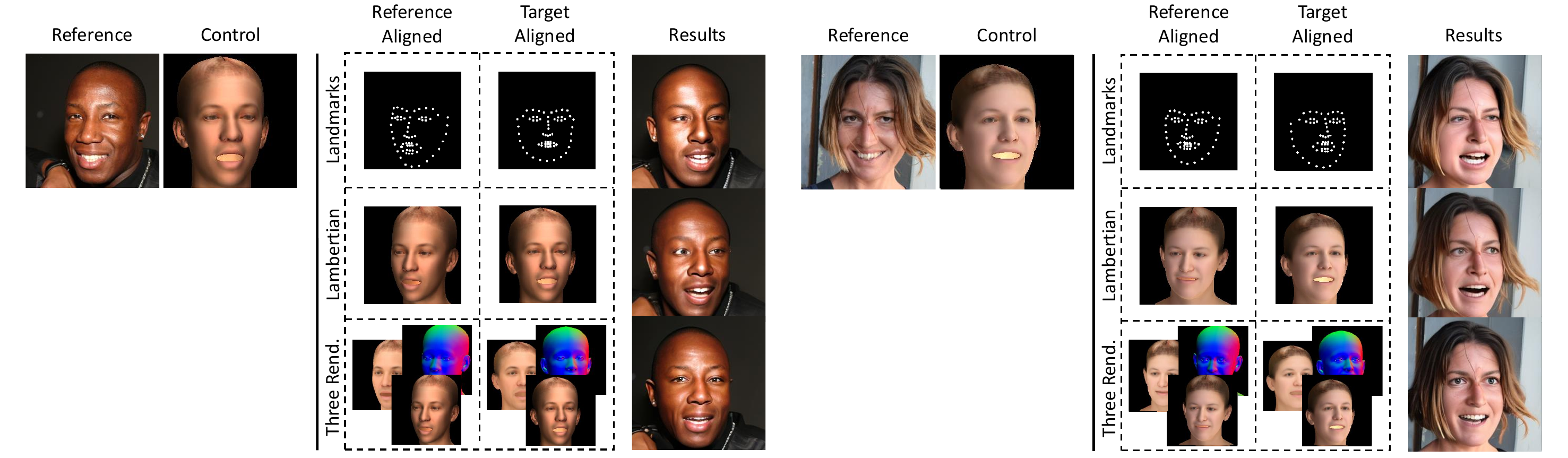}
\vspace{-15pt}
\caption{\textbf{Qualitative results of CMM input ablation.} We provide generated images on three different models which takes different inputs for CMM. }
\label{fig:lmk_qual}
\end{center}
\vspace{-20pt}
\end{figure}

The results, presented in Tab.~\ref{tab:lmk}, demonstrate that as the input provides richer information about the image, the CMM is better able to embed the relationship, resulting in improved control adherence performance. We also provide Fig.~\ref{fig:lmk_qual} to visualize the different inputs for the CMM and to compare the three models. Fig.~\ref{fig:lmk_qual} shows that the two models, which take either facial landmarks or only Lambertian rendering as CMM input, both lack control adherence compared to our method. Also, the model with landmark input compromises the identity.

We also trained three models with different architectures on CMM. Fig.~\ref{tab:cmm_arc} shows that our design of CMM exhibits the best results.

\begin{table}[t]
    \centering
    \caption{\textbf{Quantitative results of training DiffuionRig~\cite{ding2023diffusionrig} on video dataset.} We train DiffuisonRig~\cite{ding2023diffusionrig} on video dataset~\cite{zhu2022celebvhq} and evaluate following the same protocol introduced in the main paper.}
    \vspace{-2mm}

    \begin{tabular}{l@{\hspace{5pt}}|c@{\hspace{5pt}}c@{\hspace{5pt}}c@{\hspace{5pt}}c@{\hspace{5pt}}c@{\hspace{5pt}}}
    \hline
      DiffusionRig~\cite{ding2023diffusionrig}&Re-Infer.$\downarrow$&  ID $\uparrow$ &  FID $\downarrow$ &LPIPS $\downarrow$ \\
    \hline
     - FFHQ~\cite{karras2019style}
     &  5.06 & 0.2042& 23.05 & 0.3758 \\

    - CelebV-HQ~\cite{zhu2022celebvhq} (Recon.)
    &  5.78 & 0.0712 & 55.37 & 0.3889 \\

       - CelebV-HQ (Paired)
     &  5.77 & 0.0670 & 59.71 & 0.4036 \\
     \hline

    \textbf{ControlFace (Ours)}
     &  \textbf{4.85} & \textbf{0.7586} & \textbf{15.50} & \textbf{0.1429} \\
    \hline
    \end{tabular}
    % \vspace{-pt}    
    \label{tab:vid}
\end{table}

\begin{table}[t]
    \centering
    \caption{\textbf{Different inputs for CMM.} We train the model with different inputs for the CMM and measure the DECA~\cite{feng2021learning} re-inference error.}
    \vspace{-2mm}

    \begin{tabular}{l@{\hspace{5pt}}|c@{\hspace{5pt}}c@{\hspace{5pt}}c@{\hspace{5pt}}c@{\hspace{5pt}}|c@{\hspace{5pt}}c@{\hspace{5pt}}}
    \hline
      &Light $\downarrow$&  Shape $\uparrow$ &  Exp. $\downarrow$ & Pose $\downarrow$ & Avg. $\downarrow$ \\
    \hline
     Landmarks
     &  3.88 & 2.65& 6.63 & \textbf{6.63} &9.50\\

    Lambertian Rendering
    &  4.11 & \textbf{2.56}& 5.83 & 7.3 & 4.95 \\

     \hline

    \textbf{Three Renderings (Ours)}
     &  \textbf{3.75} & \textbf{2.56} & 5.43 & 7.67 & \textbf{4.85}\\
    \hline
    \end{tabular}
    \label{tab:lmk}
\end{table}

\begin{table}[t]
    \centering
    \caption{\textbf{Different architectures for CMM.} We train the model with different architectures for the CMM and measure the DECA~\cite{feng2021learning} re-inference error and identity similarity (ID)~\cite{deng2019arcface}.}
    \vspace{-2mm}

    \begin{tabular}{l|cccc}
                    \toprule
                    Variants & Re-Infer ($\downarrow$) & ID ($\uparrow$) \\
                    \midrule
                    CMM with ResNet architecture            & 5.27   & 0.7199 \\
                    No CMM, Face Controller on reference    & 5.40   & 0.7184 \\
                    CMM (Reference only)                    & 5.03   & 0.7290 \\
                    \hline
                    \textbf{CMM (Ours)}                     & \textbf{4.85}   & \textbf{0.7586} \\
                    \bottomrule
                \end{tabular}
    
    \label{tab:cmm_arc}
\end{table}

\begin{figure}[t]
\begin{center}
 \includegraphics[width=0.8\linewidth]{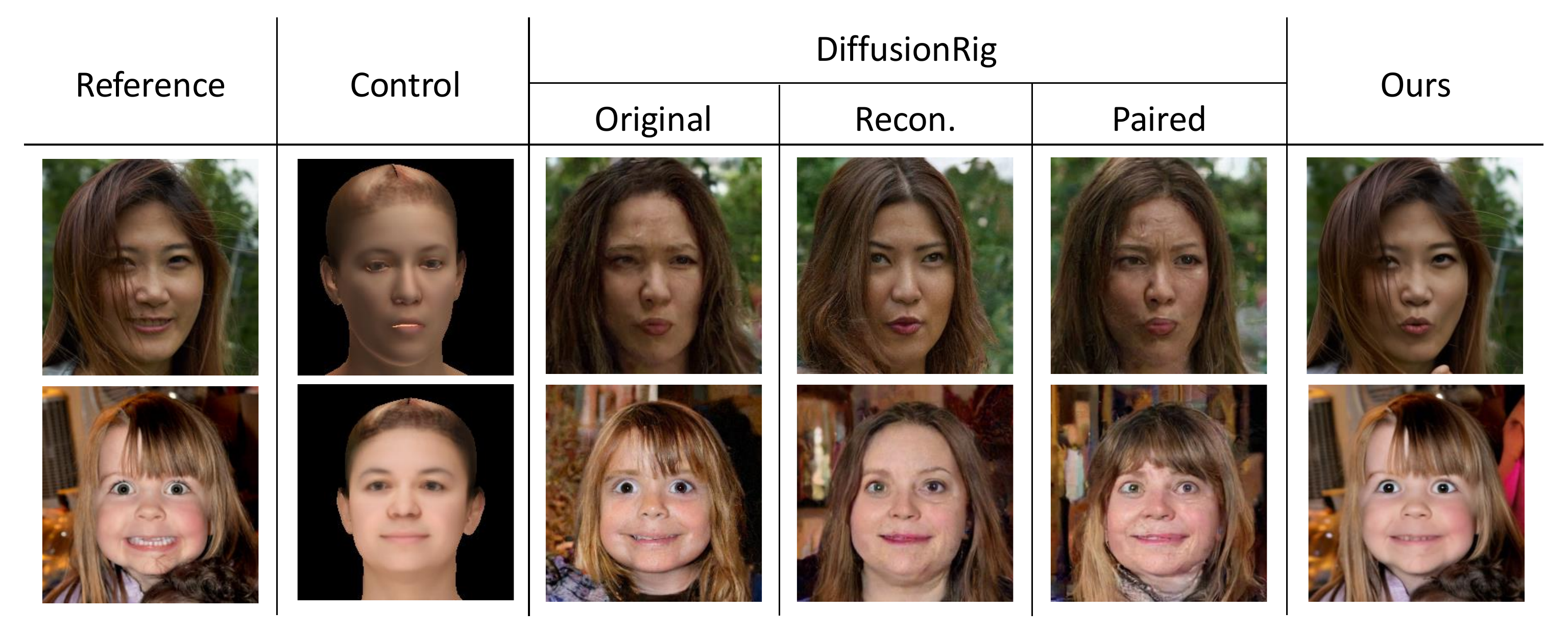}

\vspace{-5pt}
\caption{\textbf{Qualitative results of training DiffuionRig~\cite{ding2023diffusionrig} on video dataset.} We illustrate the results of two variants of DiffusionRig that is trained on CelebV-HQ~\cite{zhu2022celebvhq}. }
\label{fig:vidirig}
\end{center}

\vspace{-20pt}
\end{figure}

\begin{figure}[t]
    \centering
    % First PDF as subfigure (a)
    \begin{subfigure}[b]{0.8\textwidth} % Adjust the width (relative to text width)
        \centering
        \includegraphics[width=0.95\linewidth]{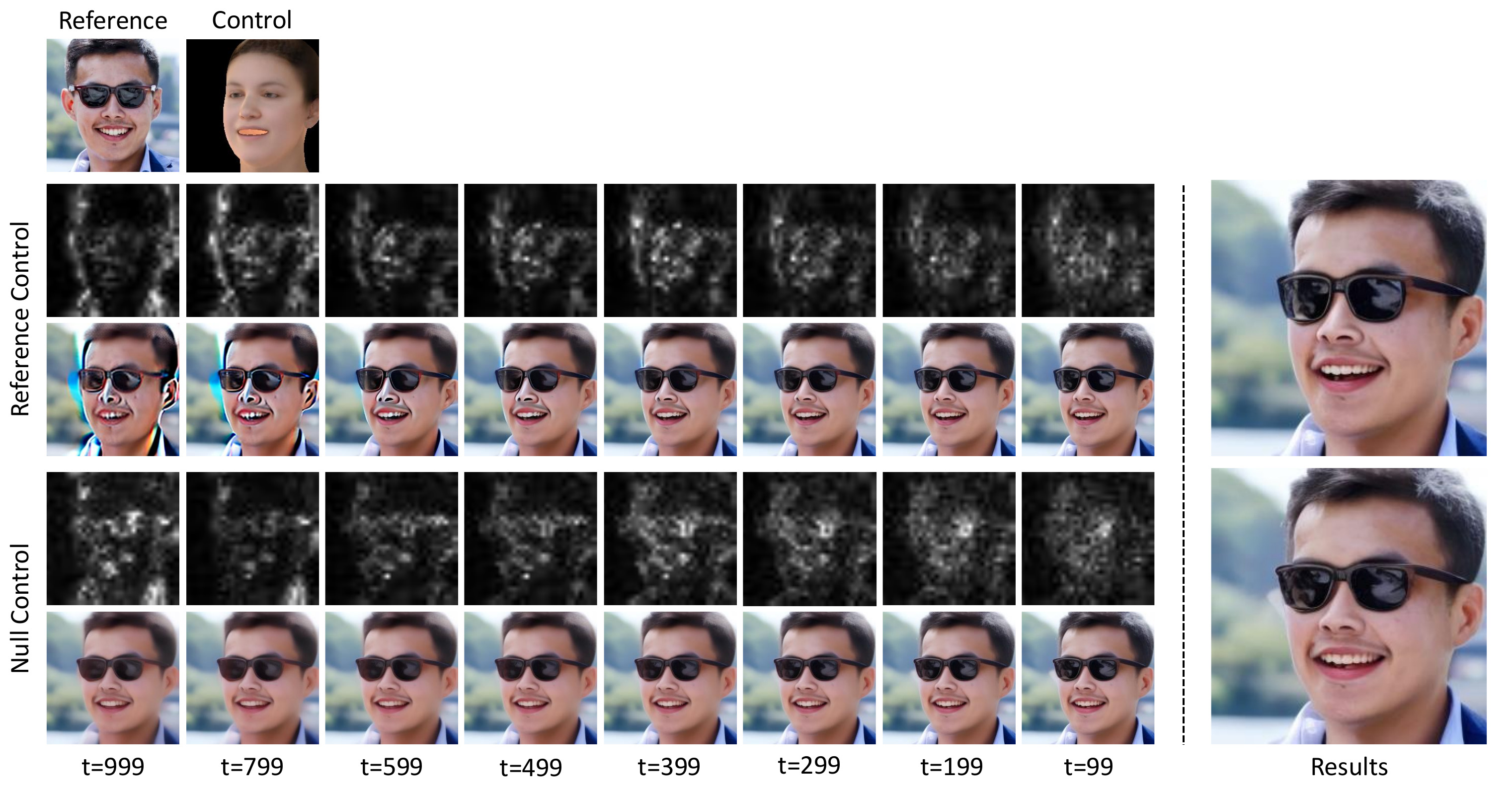} % Replace with your first PDF file
        \caption{}
        \label{fig:sub_a}
    \end{subfigure}
    \vspace{5pt}
    % Second PDF as subfigure (b)
    \begin{subfigure}[b]{0.8\textwidth} % Adjust the width
        \centering
        \includegraphics[width=0.95\linewidth]{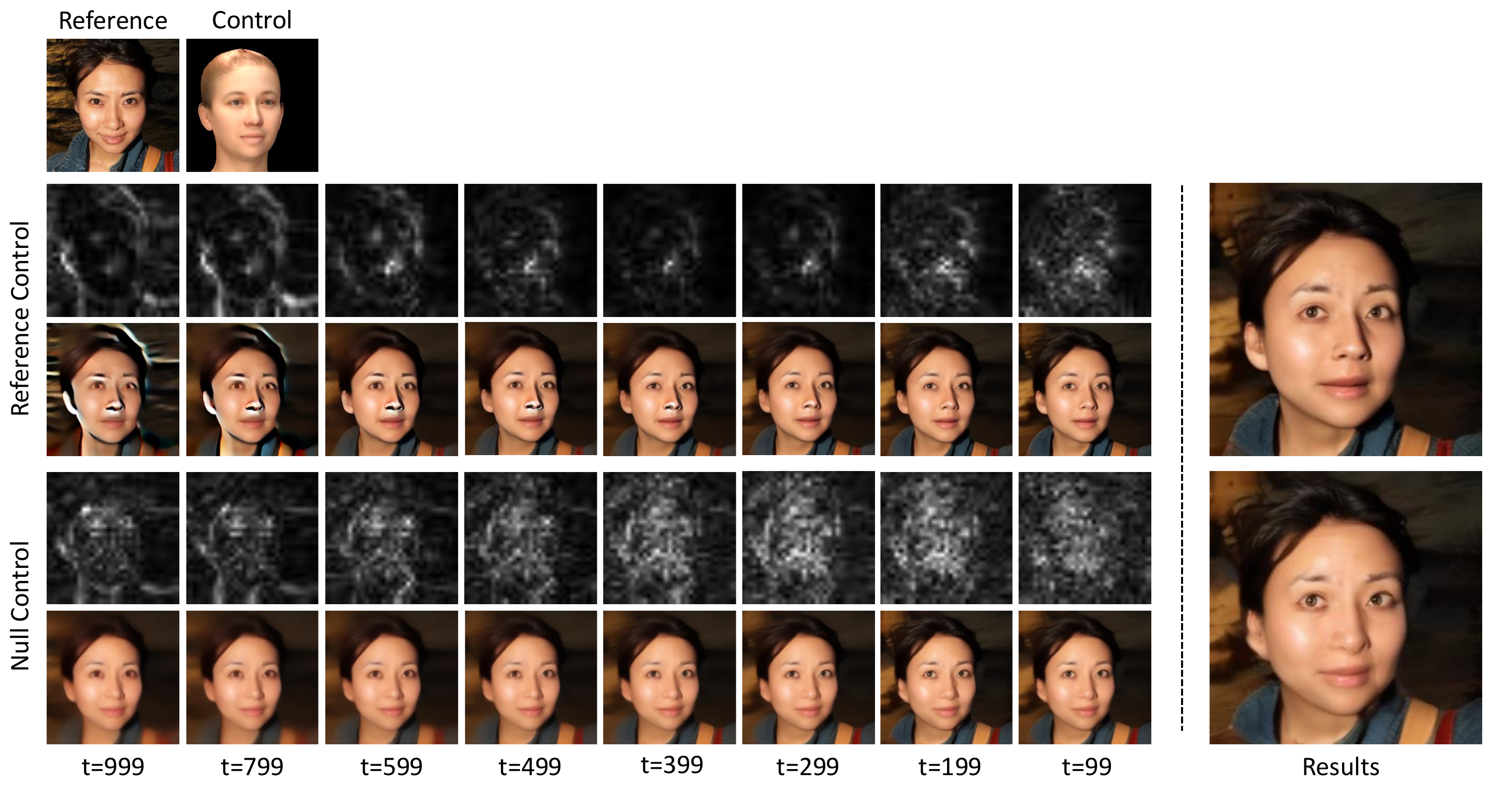} % Replace with your second PDF file
        \caption{}
        \label{fig:sub_b}
    \end{subfigure}
    \vspace{-10pt}
    \caption{\textbf{Visualization of RCG.} We visualize each difference, $ \epsilon_{\theta}(\cdot, D_{T}) - \epsilon_{\theta}(\cdot, \varnothing)$ and $ \epsilon_{\theta}(\cdot, D_{T}) - \epsilon_{\theta}(\cdot, D_{R})$, along with the predicted $X_0$ on each timesteps $t$.}\vspace{-20pt}
    \label{fig:ap_rcg}
\end{figure}

\subsection{More RCG Visualization}
\label{ap:rcg}
We provide more analysis on reference control guidance (RCG). In Fig.~\ref{fig:ap_rcg}, we visualize the differences, $ \epsilon_{\theta}(\cdot, D_{T}) - \epsilon_{\theta}(\cdot, \varnothing)$ and $ \epsilon_{\theta}(\cdot, D_{T}) - \epsilon_{\theta}(\cdot, D_{R})$ which corresponds to CFG~\cite{ho2022classifier} applied to pose controller and RCG, respectively, across various timesteps $t$. Additionally, we visualize the predicted $X_0$ to observe how quick each method converges to the final output. The second and fourth row in Fig.~\ref{fig:sub_a} and Fig.~\ref{fig:sub_b} show the differences and the third and fourth row display the predicted $X_0$. 

Interestingly, RCG exhibits over-saturated predicted $X_0$ whereas CFG generates over-smoothed predictions. This happens because RCG is better grounded, with the differences focused in specific areas where the output of the denoising U-Net varies between the reference and target controls. This over-saturation resolves quickly, resulting in realistic predictions faster than the slow dissipation of smoothness observed in CFG, which contributes to blurriness in the final generation result. As shown in Fig.~\ref{fig:sub_b}, the generation result of CFG still exhibits this blurriness both on the face and the right shoulder. Additionally, the generation results in Fig.~\ref{fig:sub_a} demonstrate that RCG achieves better adherence to the target control. RCG guides the sampling process to align more closely with the target control, where the face is oriented to the left. On the other hand, the result of CFG faces slightly forward.

\section{Additional Qualitative Results}
We provide more rigged results geenerated by ControlFace. First, we show generated results with faces acquired from FFHQ~\cite{karras2019style}. Then, we demonstrate that ControlFace can be applied to rig out-of-distribution faces such as cartoon-like portraits showing its strong robustness and generalizability. For out-of-distribution results, we utilize two types of face images: those manually collected from the internet and those generated using a large pretrained text-to-2D diffusion model.

\subsection{FFHQ}
\label{ap:qual ffhq}

\begin{figure}[h]
\centering
\includegraphics[width=1\textwidth]{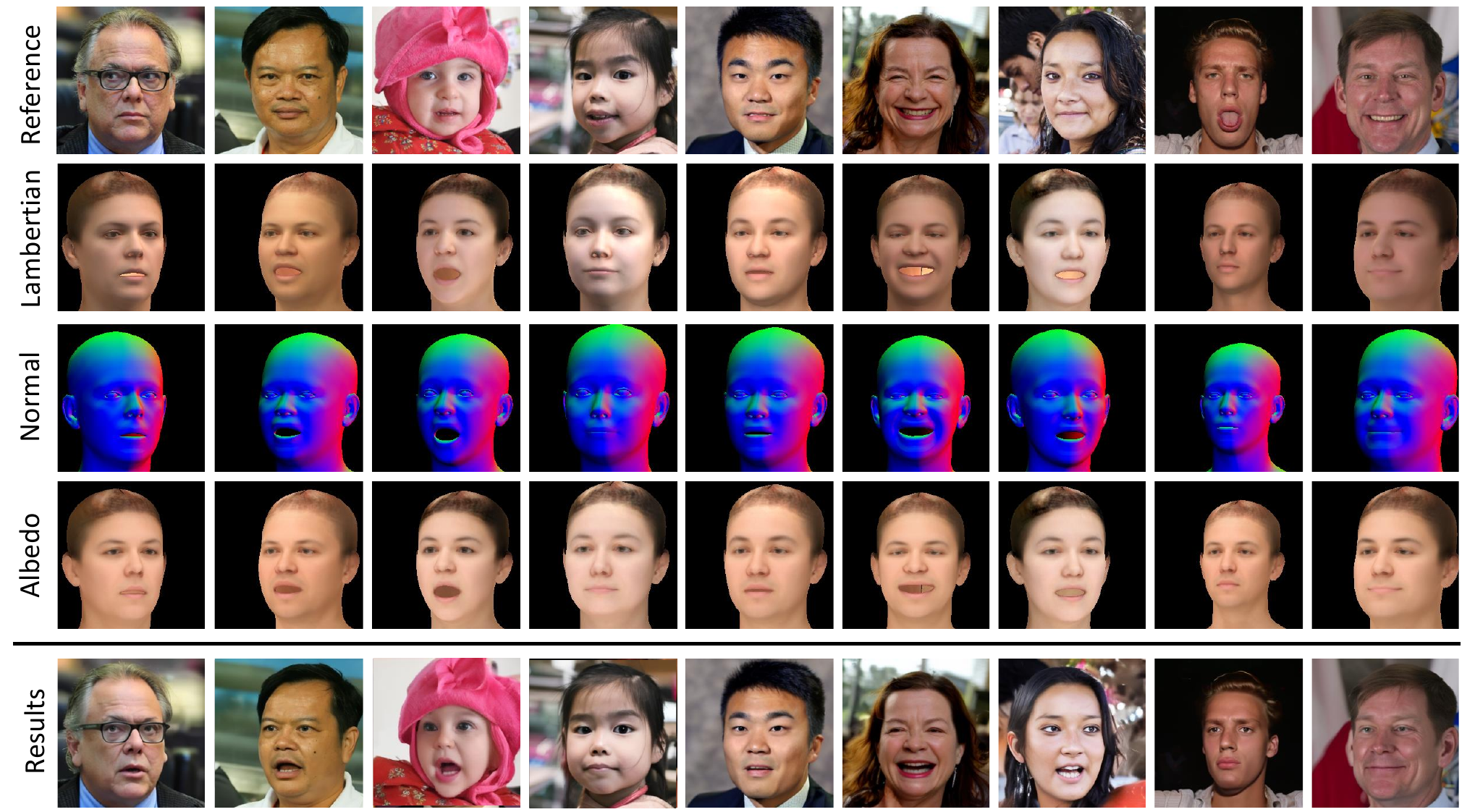}
    \caption{\textbf{Additional results on FFHQ~\cite{karras2019style}.} We randomly select faces from FFHQ~\cite{karras2019style} and rig the pose parameters.}
\vspace{50pt}
\end{figure}

\begin{figure}[h]
\includegraphics[width=\textwidth]{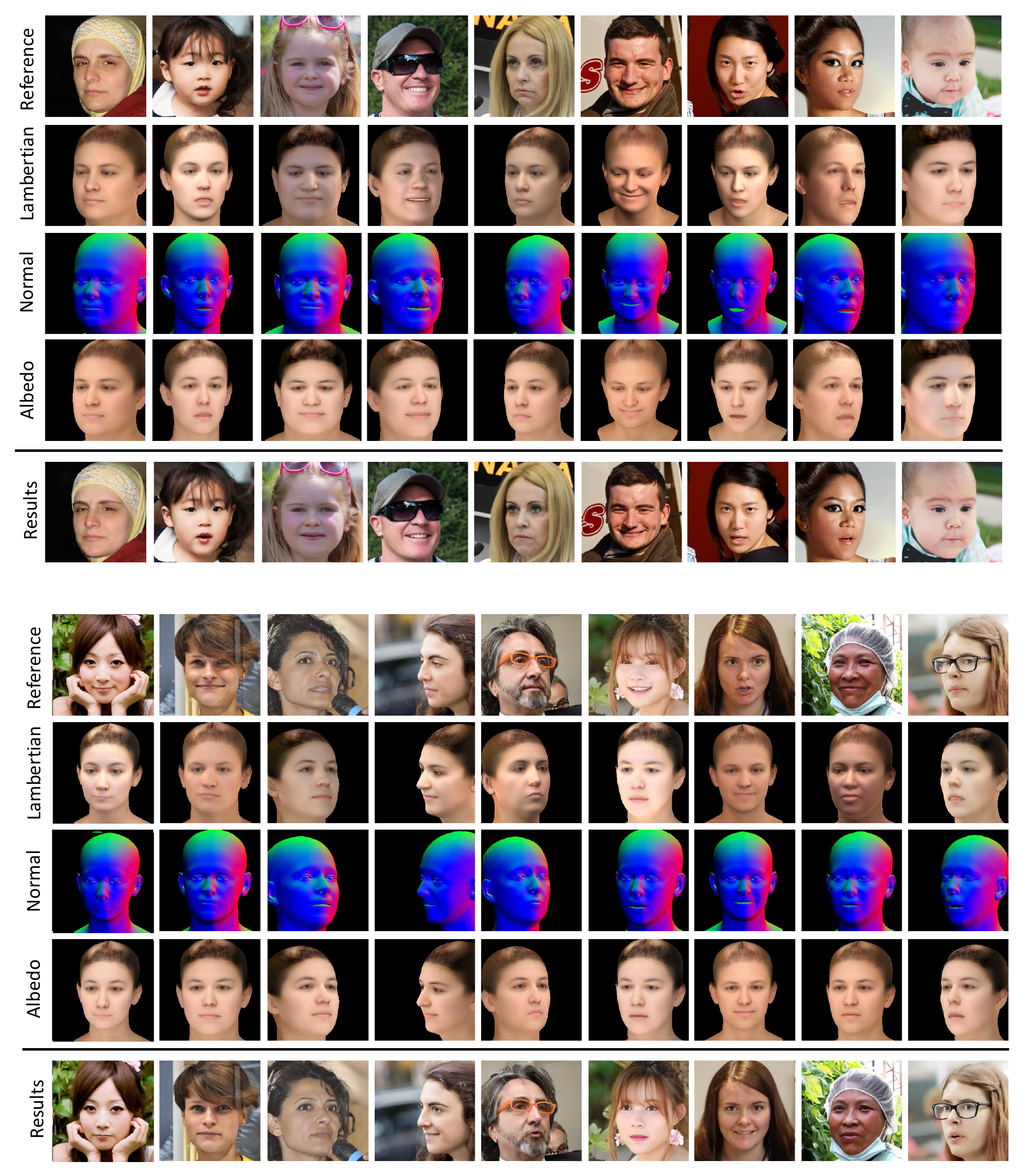}
\caption{\textbf{Additional results on FFHQ~\cite{karras2019style}.}  We randomly select faces from FFHQ~\cite{karras2019style} and rig the shape parameters for the results on the top and expression parameters for the bottom.}
\vspace{5pt}
\end{figure}

\begin{figure}[t]
\centering
\includegraphics[width=\textwidth]{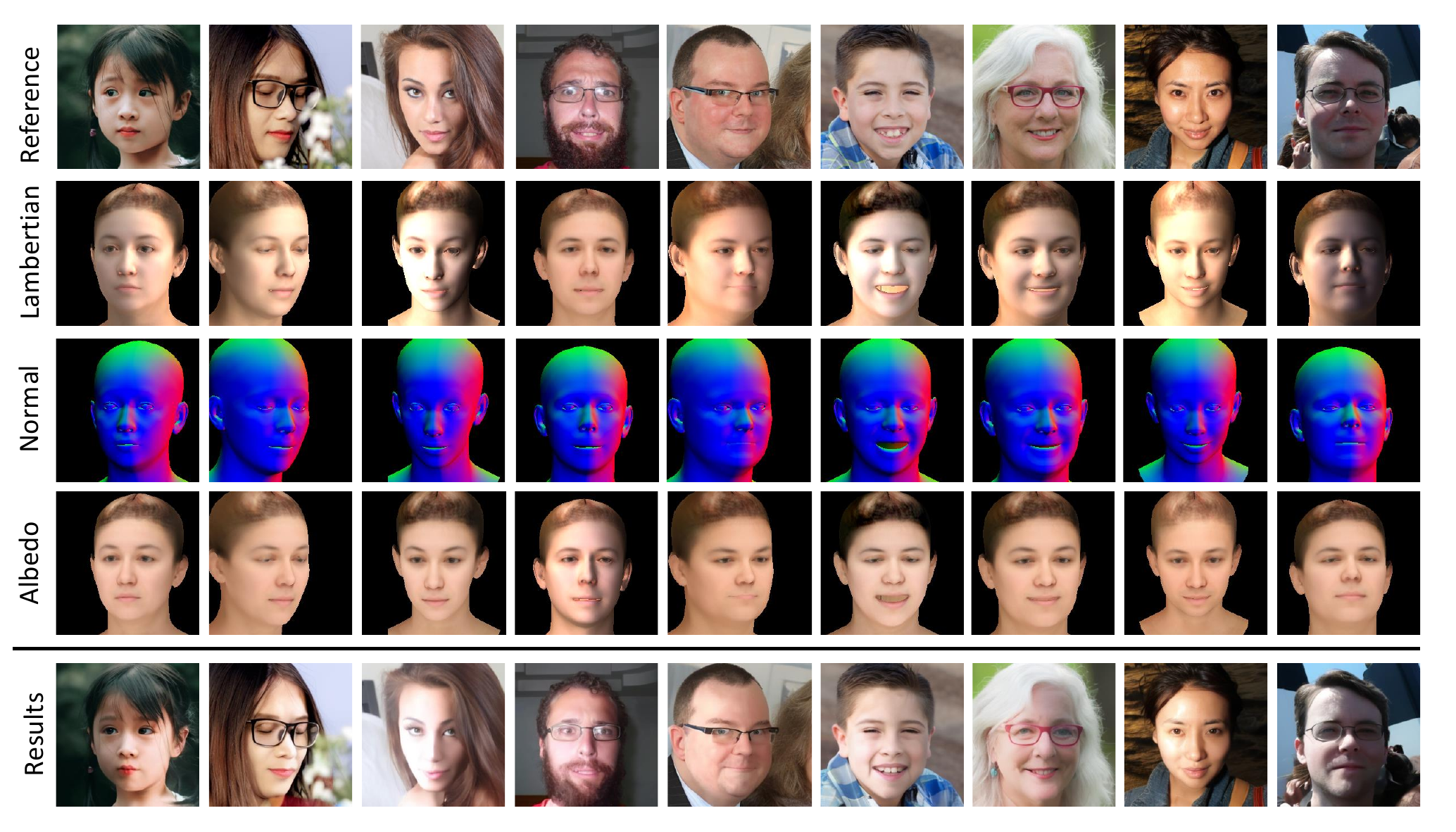}
    \vspace{-20pt}
    \caption{\textbf{Additional results on FFHQ~\cite{karras2019style}.}  We randomly select faces from FFHQ~\cite{karras2019style} and rig the light parameters. }
\vspace{-10pt}
\end{figure}

\subsection{Out-of-Distribution}
\label{ap:qual ood}
\begin{figure}[h]
\centering
\includegraphics[width=\textwidth]{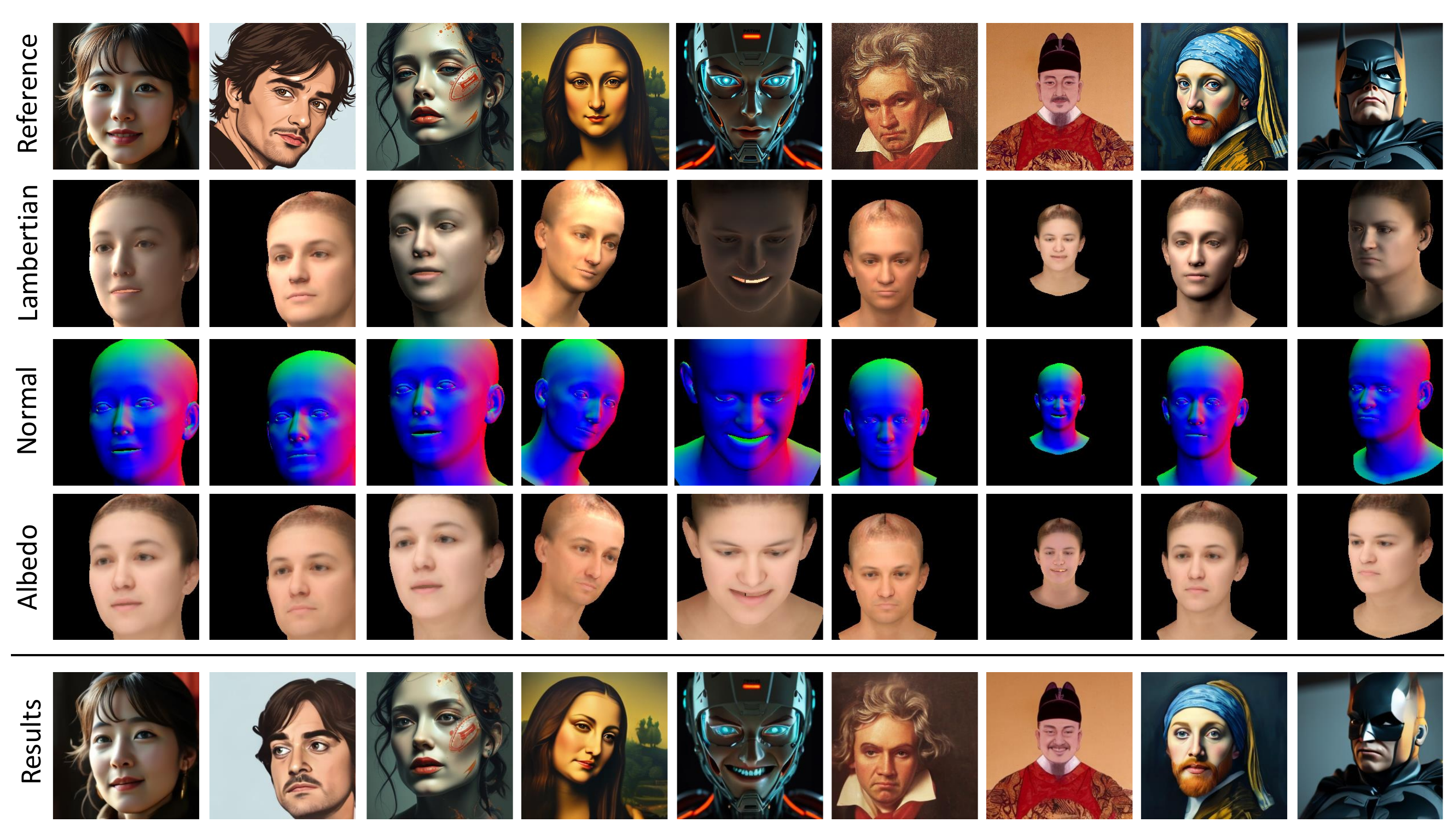}
    \vspace{-20pt}
    \caption{\textbf{Additional results on out-of-distribution faces.} We apply ControlFace on portraits obtained from the internet and generated by a text-to-2d diffusion model.}
    \vspace{-10pt}
\end{figure}

\end{document}